\documentclass[lettersize,journal]{IEEEtran}
\usepackage{amsmath,amsfonts}
\usepackage{algorithmic}
\usepackage{algorithm}
\usepackage{array}
\usepackage[caption=false,font=normalsize,labelfont=sf,textfont=sf]{subfig}
\usepackage{textcomp}
\usepackage{stfloats}
\usepackage{url}
\usepackage{verbatim}
\usepackage{graphicx}
\usepackage{cite}
\usepackage{amsmath}

\usepackage{booktabs}
\usepackage{multirow}
\usepackage{tabularx}
\usepackage{hyperref}
\hypersetup{hypertex=true,
colorlinks=true,
linkcolor=blue,
anchorcolor=blue,
citecolor=blue}
\begin{document}

\title{Take Fake as Real: Realistic-like Robust Black-box Adversarial Attack to Evade AIGC Detection}
\author{Caiyun~Xie, Dengpan~Ye$^{\ast}$,~\IEEEmembership{Member,~IEEE}, Yunming~Zhang, Long~Tang, Yunna~Lv, Jiacheng~Deng, Jiawei~Song
% <-this % stops a space
\thanks{Caiyun~Xie, Dengpan~Ye, Yunming~Zhang, Long~Tang, Yunna~Lv, Jiacheng~Deng, Jiawei~Song are with the Key Laboratory of Aerospace Information Security and Trusted Computing, Ministry of Education. School of
Cyber Science and Engineering, Wuhan University, China, 430072. (e-mail:
\{caiyunxie, yedp, zhangyunming, l\_tang, lvyunna, dengjiacheng\}@whu.edu.cn, jovesang@qq.com) }% <-this % stops a space
\thanks{*Corresponding author}
}

\markboth{}%Journal of \LaTeX\ Class Files,~Vol.~14, No.~8, August~2021
{Shell \MakeLowercase{\textit{et al.}}: A Sample Article Using IEEEtran.cls for IEEE Journals}


\maketitle
\begin{abstract}
The security of AI-generated content (AIGC) detection is crucial for ensuring multimedia content credibility. To enhance detector security, research on adversarial attacks has become essential. However, most existing adversarial attacks focus only on GAN-generated facial images detection, struggle to be effective on multi-class natural images and diffusion-based detectors, and exhibit poor invisibility. To fill this gap, we first conduct an in-depth analysis of the vulnerability of AIGC detectors and discover the feature that detectors vary in vulnerability to different post-processing. Then, considering that the detector is agnostic in real-world scenarios and given this discovery, we propose a Realistic-like Robust Black-box Adversarial attack (R$^2$BA) with post-processing fusion optimization. Unlike typical perturbations, R$^2$BA uses real-world post-processing, i.e., Gaussian blur, JPEG compression, Gaussian noise and light spot to generate adversarial examples. Specifically, we use a stochastic particle swarm algorithm with inertia decay to optimize post-processing fusion intensity and explore the detector's decision boundary. Guided by the detector's fake probability, R$^2$BA enhances/weakens the detector-vulnerable/detector-robust post-processing intensity to strike a balance between adversariality and invisibility. Extensive experiments on popular/commercial AIGC detectors and datasets demonstrate that R$^2$BA exhibits impressive anti-detection performance, excellent invisibility, and strong robustness in GAN-based and diffusion-based cases. Compared to state-of-the-art white-box and black-box attacks, R$^2$BA shows significant improvements of 15\%--72\% and 21\%--47\% in anti-detection performance under the original and robust scenario respectively, offering valuable insights for the security of AIGC detection in real-world applications.
\end{abstract}

\begin{IEEEkeywords}
Adversarial attack, AIGC detection, Image processing, Particle swarm optimization
\end{IEEEkeywords}

\section{Introduction}
\IEEEPARstart{W}{ith} the rapid development of GAN- and diffusion-based generation technologies\cite{Diffusion_privacy}, AI-generated content (AIGC) detection has become a popular research topic\cite{DIRE},\cite{GenImage},\cite{DRCT}. The security of state-of-the-art (SOTA) AIGC detection is crucial for ensuring multimedia content credibility\cite{security_23}. Without secure AIGC detection technologies, multimedia content could face a serious trust crisis. However, AIGC detection is still in an immature stages: Many studies\cite{AdvThreats},\cite{AdvDeepfake},\cite{Exploring},\cite{Evade} have shown that GAN-based detectors are vulnerable to adversarial attacks, leading to detection failures. On the other hand, the widespread use of diffusion models makes security research on this detection method more urgent. However, the lack of robustness research\cite{DIRE},\cite{DRCT},\cite{UnivFD},\cite{FatFormer},\cite{ConvB}, combined with the distinct generative mechanisms of diffusion models compared to GANs\cite{DDIM},\cite{DM_Survey}, leaves the security of diffusion-based AIGC detection uncertain. Therefore, to improve the security of AIGC detection, it is crucial to focus on the research of these technologies' robustness against adversarial attacks, particularly for diffusion-based detectors.

\begin{figure}[!t]
\centering
\includegraphics[width=3.48in]{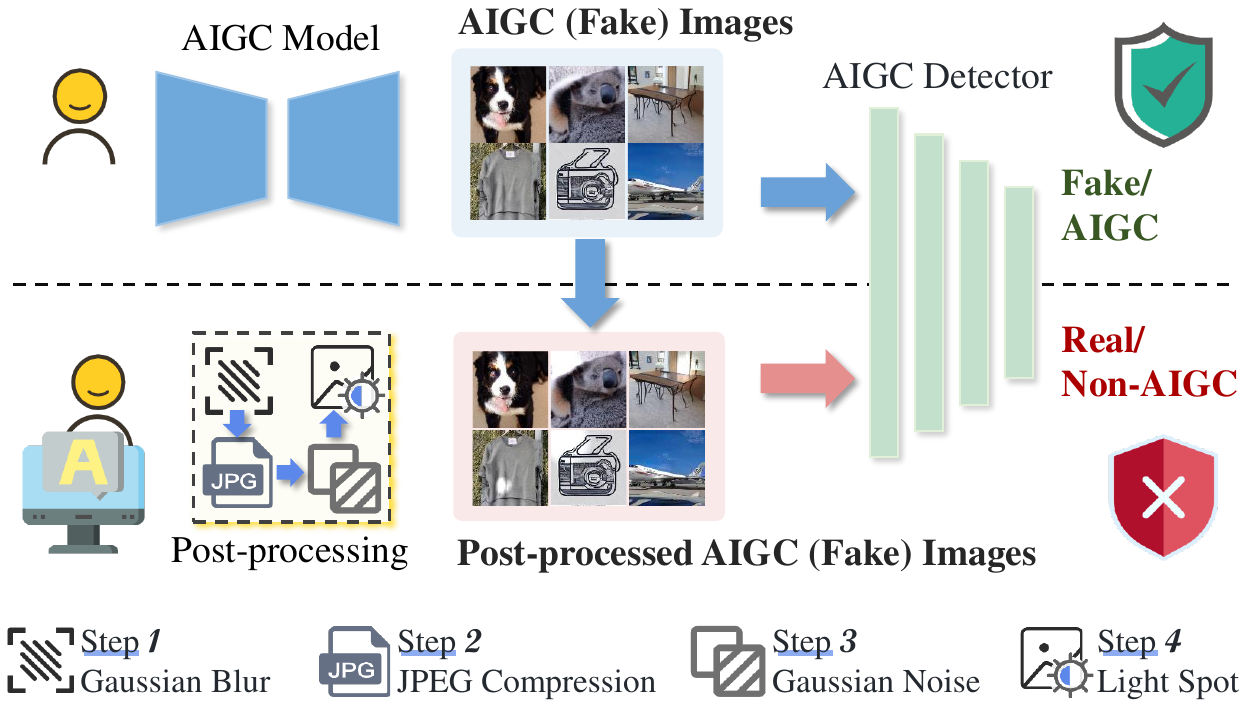}
\caption{Principle of our method. We use Gaussian blur, JPEG compression, Gaussian noise and light spot to process AI generated images. The post-processed image serves as an adversarial example that leads the detector to recognize the AI-generated image as a real/non-AI-generated image.}
\label{fig:intro}
\end{figure}

The existing adversarial attacks against AIGC detectors primarily focus on GAN-based detectors. Some studies\cite{AdvThreats},\cite{AdvDeepfake},\cite{Exploring},\cite{AVA_SP} employ gradient-based attacks to generate perturbations, but these perturbations are always recognized as fake by well-trained detectors. Additionally, these studies\cite{AdvThreats},\cite{AdvDeepfake},\cite{Exploring},\cite{Evade},\cite{AVA_SP} perform well on Deepfake facial datasets but poorly on multi-class natural image datasets, which restricts their broader application. Furthermore, the invisibility of existing adversarial examples\cite{Evade},\cite{BruSLe} is often overlooked, leading to severe image distortions that limit their practical use. Therefore, designing realistic-like adversarial attacks applicable to multi-class natural images is of significant practical value for deploying AIGC detectors in real-world scenarios. 

Previous researches\cite{Evade},\cite{Adversarial_Relighting},\cite{AVA},\cite{Can2022},\cite{WTEvade} demonstrate that adversarial examples can be generated by employing natural images degradation as an attack method. These methods generate adversarial examples by simulating the degradation of image quality in the real world, which is more meaningful for studying the robustness of AIGC detectors, as post-processing techniques, like JPEG compression and Gaussian blur, are common on social networks. In other words, if a detector classifies such post-processed AIGC images as real, it suggests that the detector will face similar performance degradation issues in real-world applications. Therefore, employing post-processing techniques to generate adversarial samples is both feasible and of significant practical value. 

In order to design a realistic-like adversarial attack applicable to natural images, as shown in Fig. \ref{fig:intro}, we select Gaussian blur, JPEG compression, Gaussian noise, and light spot as attacks to generate adversarial examples. These 4 techniques have been proven to attack the frequency, spatial, and statistical domains of detectors, respectively\cite{Evade},\cite{Adversarial_Relighting},\cite{Can2022},\cite{WTEvade}. And the pre-experiment we conduct in Fig. \ref{fig:proof} reveals the discovery that detectors vary in vulnerability to different post-processing. Based on the discovery, and considering that the detector is agnostic in real-world scenarios, we propose R$^2$BA, a realistic-like robust black-box adversarial attack based on post-processing fusion optimization. During the optimization process, R$^2$BA first randomly generates multiple post-processed adversarial examples to explore the detector's complex decision boundary. Then, guided by the fake probability output of the AIGC detector, R$^2$BA continuously updates the post-processing intensity of these adversarial examples, driving them toward and eventually across the decision boundary of the detector, thereby evading AIGC detection.

R$^2$BA solves 3 problems that existed in previous works:

{\textbf{Applicability to AIGC detectors.}} Based on the discovery that detectors vary in vulnerability to different post-processing techniques, we integrate 4 widely-used post-processing techniques to launch a comprehensive attack, ensuring all detectors are successfully attacked by R$^2$BA, resulting in significant detection performance degradation. 

{\textbf{Balance between invisibility and adversariality.}} Guided by the detector's fake probability, R$^2$BA enhances the detector-vulnerable post-processing intensity to improve adversariality, and weakens the detector-robust post-processing intensity to minimize unnecessary modifications, making the post-processed AIGC images appear highly invisible and realistic from a human visual perspective.

{\textbf{Multi-class natural images effectiveness.}} 
R$^2$BA is not a class-wise method and does not set any constraints on data distribution. To address the complex decision boundary issues introduced by multi-class natural images, R$^2$BA randomly initialize multiple post-processed images to expand the search space. This approach allows us to efficiently locate the decision boundaries that are most vulnerable to the fusion post-processing, thus evading detection. 

Compared to existing adversarial attacks against AIGC detectors, R$^2$BA generates adversarial examples with real-world significance. It demonstrates powerful anti-detection performance on AIGC detectors for natural images, not limited to facial images or GAN-based detectors, thereby filling the gap in adversarial attacks against both GAN-based and diffusion-based AIGC detectors.

Our contributions can be summarized as follows:
\begin{itemize}
\item{We propose R$^2$BA, a novel realistic-like robust black-box adversarial attack. As the first method to evade AIGC detectors on natural images, R$^2$BA provides significant practical value. It demonstrates applicability to black-box detectors, with impressive anti-detection performance, excellent image quality, and remarkable robustness.}
\item{We innovatively propose a post-processing fusion optimization method, which adjusts post-processing intensity to help post-processed AIGC images evade detection. Guided by the detector's vulnerability to different post-processing, we use a stochastic particle swarm optimization algorithm with inertia decay to address the complex trade-off between adversariality and invisibility.}
\item{We conduct extensive experiments on the CNNSpot, GenImage, and DRCT datasets to validate the effective anti-detection performance of R$^2$BA in evading SOTA GAN-based and diffusion-based AIGC detectors, including the commercial AIGC detector from Alibaba. Compared to existing attacks, R$^2$BA achieves average improvements of 15\%--72\%, 21\%--47\%, and 26\%--68\% in adversariality, robustness and invisibility, respectively.}
\end{itemize}

The remainder of this paper is organized as follows: Section \ref{section:RelatedWork} reviews related work. Section \ref{section:ProposedMethod} presents the specific implementation of the proposed method. The experiments and analyses are described in Section \ref{section:Experiments}. Finally, we provide a brief conclusion in Section \ref{section:Conclusion}.

\section{Related Work}\label{section:RelatedWork}
\subsection{AI-generated Image Forgery Detection}
AI-generated image forgery detection is actually a {\slshape{Fake-Real}} binary classification problem. Here, {\slshape{Fake}} refers to AIGC, which means the image is generated by AI, specifically through techniques like GANs or diffusion models. While {\slshape{Real}} refers to UGC, meaning that the image is a natural image or user-generated. The challenge of AIGC detection lies in extracting subtle forgery signals from AI-generated images and making accurate distinctions. According to different generation technologies, AIGC detection techniques can be divided into GAN-based and diffusion-based methods.

{\textbf{GAN-based forgery detection.}} 
Image generation based on GAN architecture\cite{StyleGAN},\cite{ProGAN},\cite{BigGAN} consists of generative and discriminative models. The goal is to train the generative model to match the real data distribution, making it difficult for the discriminative model to distinguish between real and AI-generated images. To prevent misuse, several GAN-based forgery detection techniques\cite{UnivFD},\cite{FatFormer},\cite{CNNSpot},\cite{ResNetND},\cite{DeepFakeDetection} have been proposed. These methods detect AI-generated images by extracting specific features and using data-driven approaches to learn differences across spatial, frequency, and statistical domains between AI-generated and natural images.

{\textbf{Diffusion-based forgery detection.}} 
Diffusion-based image generation\cite{LDM},\cite{VQDM},\cite{Glide} has gained considerable attention due to its powerful generative capabilities. Unlike the GAN architecture, diffusion models completely revolutionize the traditional generative process. These models use high-dimensional latent variables to capture more details of the data, resulting in higher-quality images. It is evident that the powerful generative ability of diffusion models enhances the robustness of diffusion-based forgery detection techniques\cite{DIRE},\cite{DRCT},\cite{UnivFD},\cite{FatFormer}. These techniques focus on capturing subtle forgery signals in diffusion-generated images through methods such as reconstruction or contrast learning, aiming to identify small differences that may be exist in the image, and thus accurately distinguishing between real and fake content.

\subsection{Adversarial Attacks for Anti-detection}
In order to explore the robustness of detectors, many adversarial attacks have been used to generate adversarial examples for anti-detection. These works aim to generate adversarial examples for AI-generated images, which cause the detector to misclassify the fake/AI-generated images as real images. However, current anti-detection adversarial attacks\cite{AdvThreats},\cite{AdvDeepfake},\cite{Evade} are mainly performed on the Deepfake face dataset, which limits their broader applications and often results in severe performance degradation when against robust AIGC detectors on the multi-class AI-generated natural image datasets. The following are some popular adversarial attack methods.

{\textbf{White-box attack.}} 
Classical white-box attack methods\cite{AdvThreats},\cite{AdvDeepfake},\cite{Exploring} rely on obtaining gradient information from the target model to generate adversarial examples. But the meaninglessness and randomness of these perturbations may make such algorithms fail to attack in more complex tasks or scenarios. Some studies\cite{Evade},\cite{AVA_SP} aim to evade GAN-based detection through statistical feature consistency or inconspicuous attribute variation in facial images. However, these methods often fail because it is difficult to determine the features that the detector truly focuses on, and the attributes of natural images are highly diverse, with significant differences across different classes of images. 

{\textbf{Black-box attack.}} Black-box adversarial attacks do not require access to the inner workings of the model, and efficient attacks can be achieved by only obtaining the model output. There are three main types of black-box attacks: Algorithms based on gradient estimation\cite{AdvDeepfake} and agent models\cite{MGAA} are limited due to their reliance on fitting the target. When applied in practice, query-based methods might be the best solution. However, when dealing with the unpredictable decision boundaries caused by multi-class natural images\cite{AdversarialDecision}, locating the decision boundary for a single adversarial example requires considerable effort. Some methods based on random pixel updates\cite{BruSLe},\cite{Square} are effective in evading detection, but tend to introduce significant perturbations, which causes these adversarial examples to be directly classified as forged images by the human eye, even without a detector. Another method involving attribute editing\cite{AVA_SP}, although optimizing multiple adversarial examples simultaneously, still face the challenge of selecting the appropriate attribute across the diverse natural images and requires suitable attribute editing tools.

Based on the above analysis, existing adversarial attacks for anti-detection have limitations in image classes (they only work on facial datasets) and detectors (primarily GAN-based detection). Additionally, they are often difficult to strike a balance between the anti-detection effectiveness and image quality. Moreover, in practical applications, real-world platform image post-processing may erase the adversarial perturbations, leading to the failure of attacks. To address these issues in existing works, we conduct comprehensive experiments on the security of GAN-based and diffusion-based AIGC detectors using natural image datasets with multiple classes, and thoroughly analyze and compare various anti-detection performance metrics of our method with existing methods in the original, robust and real-world scenarios.

\begin{figure}[!t]
\centering
\includegraphics[width=3.48in]{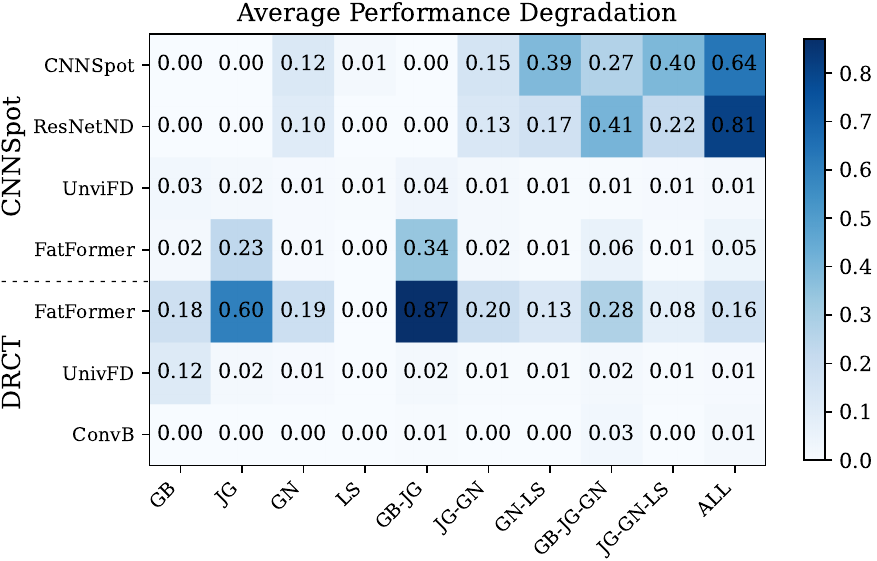}
\caption{Average performance degradation. The values represent the attack success rate of different post-processing against AIGC detectors. Darker colors indicate greater vulnerability of detectors. Attacks used include Gaussian blur (GB), JPEG compression (JG), Gaussian noise (GN), light spot (LS), and sequential fusion attacks (ALL means GB-JG-GN-LS). }
\label{fig:proof}
\end{figure}
\section{Proposed Method}\label{section:ProposedMethod}
In this section, we first explain the {\textbf{motivation}} for R$^2$BA in subsection \ref{subsection:Motivation}. Next, we use a {\textbf{visualization}} in subsection \ref{subsection:Visualization} to illustrate how R$^2$BA generates adversarial examples by optimizing post-processing intensity. Finally, we provide a detailed description of the R$^2$BA {\textbf{implementation}} in subsection \ref{subsection:Algorithm}, including specific formulas and the algorithm. 

\subsection{Motivation}\label{subsection:Motivation}
Traditional adversarial examples\cite{AdvThreats},\cite{AdvDeepfake},\cite{BruSLe},\cite{Square} generated by gradient-based or random pixel modification attacks, are both meaningless and ineffective because their poor invisibility makes them difficult to apply in real-world scenarios. They can be easily identified as fake by the human eye, without the need for a detector. In contrast, adversarial examples created using real-world post-processing techniques appear realistic and match real natural image patterns. Furthermore, if a detector classifies such post-processed AIGC images as real, it suggests that the detector will experience similar performance degradation in real-world applications.

Inspired by previous works\cite{Evade},\cite{AVA},\cite{WTEvade}, we select Gaussian blur, JPEG compression, Gaussian noise and light spot as attacks, which have been proven to mislead detectors through frequency, spatial, and statistical domains respectively. The pre-experiment we conduct in Fig. \ref{fig:proof} further confirms that these post-processing methods lead to varying degrees of performance degradation in detectors. We use Gaussian blur with a radius of 5 and deviation of 1.5, JPEG compression with an intensity of 80, Gaussian noise with standard deviation of 7/255, and light spot with an intensity of 1.8 and random radius. The datasets\cite{DRCT},\cite{CNNSpot} and detectors\cite{DRCT},\cite{UnivFD},\cite{FatFormer},\cite{CNNSpot},\cite{ResNetND} used in Fig. \ref{fig:proof} are described in Section \ref{section:Experiments}.

The results in Fig. \ref{fig:proof} demonstrate that 
{\textbf{detectors vary in vulnerability to different post-processing}}. For example, on the CNNSpot\cite{CNNSpot} dataset, the CNNSpot\cite{CNNSpot} detector is highly vulnerable to Gaussian noise but more robust to JPEG compression (GN=0.12, JG=0.00), while the FatFormer\cite{FatFormer} detector exhibits the opposite behavior (GN=0.01, JG=0.23). In addition, the pre-experiment exhibits a counterintuitive result: The effectiveness of post-processing fusion attacks does not always exceed that of individual post-processing on different detectors. On the DRCT\cite{DRCT} dataset, the attack performance of Gaussian blur + JPEG compression on the FatFormer\cite{FatFormer} is better than that of Gaussian blur alone (GB+JG=0.87, GB=0.18), whereas the opposite (GB+JG=0.02, GB=0.12) is true for the UnivFD\cite{DRCT}. As a result, optimizing the post-processing fusion intensity is challenging, but necessary.

Therefore, considering that the detector is agnostic in real-world scenarios, we adopt a black-box attack to generate adversarial examples. To attack different types of detectors, we fuse these four post-processing and define the attack order as step 1-4 in Fig. \ref{fig:intro}. To solve the challenge of optimizing the fusion intensity, we use the detector's fake probability as guidance and apply stochastic particle swarm optimization (PSO)\cite{PSO} with inertia decay to optimize the fusion intensity. Compared to other optimization algorithms, such as evolutionary algorithms and simulated annealing, stochastic PSO with inertia decay outperforms them in effectively conducting global searches in high-dimensional spaces\cite{PSORV},\cite{PSO_Black}. This method can quickly locate the global decision boundary and simultaneously optimize fusion intensity to balance invisibility and adversariality. Based on the above-mentioned, we propose a post-processing fusion optimization-based, realistic-like robust black-box adversarial attack method, R$^2$BA, to study the security of AIGC detectors in real-world applications. 

\subsection{Visualization}\label{subsection:Visualization}
\begin{figure*}[!t]
\centering
\includegraphics[width=7in]{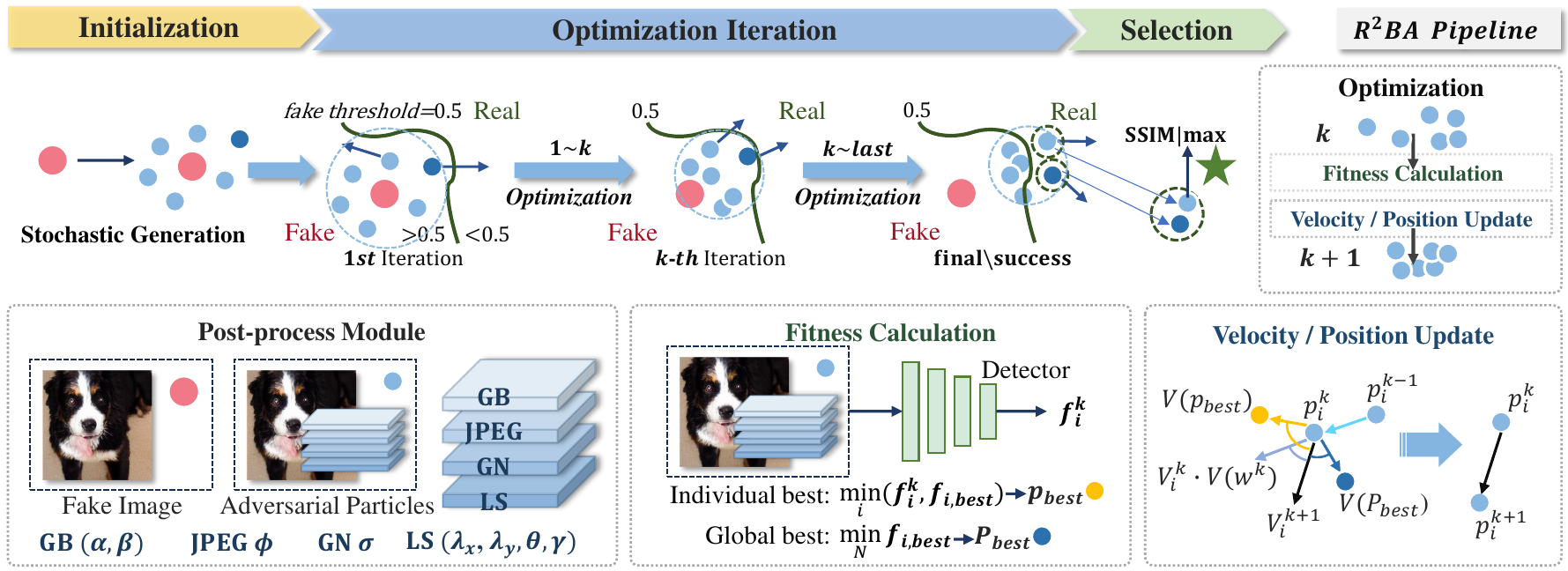}
\caption{The pipeline of R$^2$BA. It contains four modules and a main optimization process (Initialization-Optimization iteration-Selection). R$^2$BA uses \(N\) randomly generated post-processing parameters as particle positions, optimizing their velocities and positions to balance the invisibility and adversariality. A successful attack occurs when a particle crosses the detector's decision boundary, and R$^2$BA outputs the adversarial example with the highest SSIM value among the successful ones. The definitions of the notations here are the same as those in Table \ref{tab:notations}.}
\label{fig:PSO_visual}
\end{figure*}
{\textbf{Component description.}} In Fig. \ref{fig:PSO_visual}, we illustrate the pipeline of R$^2$BA. It mainly consists of four modules: the post-processing module, fitness calculation, velocity/position update, and the swarm optimization module. The decision boundary here refers to the fake threshold (0.5) of the detector, which typically classifies an image with a fake probability greater than 0.5 as fake, and less than 0.5 as real. 

\textit{1) Post-processing Module:} The red circle represents an AIGC image, and the blue circles represent its post-processed images (adversarial particles). The particle positions represent the post-processing fusion intensity. We can balance invisibility and adversariality by optimizing the particle positions. 

\textit{2) Fitness Calculation:} The post-processed image is fed into the detector to obtain its fake probability, which serves as the particle's fitness. A smaller fitness value indicates a more effective adversarial example. The individual best of each particle corresponds to the smallest fitness in the optimization process, i.e., the lowest fake probability achieved so far. The global best is the particle with the smallest fitness among all individual bests. The global best particle is most likely to cross the decision boundary, as it has the lowest fake probability.

\textit{3) Velocity / Position Update:} Velocity \(V_{i}^{k+1}\) updates are influenced by the particle's individual best \(p_{best}\), global best \(P_{best}\), and inertia \(w^{k}\). While the position of \(p_{i}^{k+1}\) updates depend on the current position of \(p_{i}^{k}\) and velocity \(V_{i}^{k+1}\). Obviously, influenced by the individual best and global best, the particles will move towards positions with lower fake probability, continuously approaching the decision boundary. 

\textit{4) Swarm Optimization Module:} To optimize the particles' positions (post-processing intensity) and reduce their fake probability, all particles should undergo fitness calculation as well as velocity and position updates.

{\textbf{Optimization Iteration.}} In the first iteration, to cover as much of the example domain space as possible, R$^2$BA randomly assigns \(N\) post-processing fusion intensity values as the initial positions of \(N\) particles. Stochastic generation helps quickly locate the detector's most vulnerable decision boundary, enabling efficient global search and addressing the challenges posed by diverse decision boundaries in multi-class natural images. In subsequent iterations, taking the $k$-th iteration as an example, guided by fitness, the particle velocity gradually points towards the decision boundary, with the swarm becoming more concentrated. In the final (or successful) iteration, some particles cross the boundary, causing the detector to misclassify them as real images. R$^2$BA then outputs the adversarial example with the highest SSIM among all the particles with successful attacks. 

{\textbf{Special designs.}} It is clear that the optimization process largely depends on whether the particles can locate the detector's decision boundary to execute the attack. To improve search capacity, we introduce inertia decay and random velocity updates to expand the search space and avoid local optima. 

The description above outlines a typical iterative process. However, in practice, AIGC images may be located near the decision boundary, causing the attack to succeed in the first iteration. If the images are far from the decision boundary, the attack may fail to cross it within the range defined by the post-processing intensity settings. Experiments show that the former case is more common, while the latter is relatively rare. 

\subsection{Implementation}\label{subsection:Algorithm}
This subsection provides the specific implementation of R$^2$BA, with the pseudocode presented in Algorithm \ref{alg:Method}. The main notations are summarized in Table \ref{tab:notations}.
\begin{table}
\renewcommand\arraystretch{1.2}
\caption{Definitions of Main Notations\label{tab:notations}}
\begin{center}
\begin{tabular}{p{1.6cm}<{\centering} p{6.3cm}<{\centering}}
\toprule  
\textbf{Notations}& \textbf{Definition}\\
\midrule
\(D\)$(\cdot)$& Victim detector with output as fake probability\\
\(N\)& The number of particles \\
\(K\)& The max number of iterations\\
\(P^k\)& The $k$-th iteration of the particle swarm\\
\(w^k\)& The inertia of the $k$-th iteration\\
\([W_{\text{min}}, W_{\text{max}}]\)& The range of inertia\\
\(C_{\text{p}}\), \(C_{\text{total}}\)& Individual confidence and particle swarm confidence\\
\(Pos\)$(\cdot)$& The position of a particle\\
\(Pos_{\text{best}}\)$(\cdot)$& Optimal position of one particle or particles\\
\(f_i^k\)& Fitness of the $i$-st particle of the $k$-st iteration\\
\(f_{i,\text{best}}\), \(f_{\text{best}}\)& Optimal fitness of one particle / particles\\
\(\alpha, \beta\)& The radius and variance of Gaussian blur\\
\(\phi\)& The compressive strength of JPEG compression\\	
\(\sigma\)& The variance of Gaussian noise\\
\((\lambda_{x},\lambda_{y})\)& The coordinates of light spot center\\
\(\theta, \gamma\)& The center brightness and  radius of light spot\\
\bottomrule
\end{tabular}
\end{center}
\end{table}

{\textbf{Particle swarm initialization.}} First, the algorithm initializes the global parameters \(N\), \(K\), \(W_{\text{min}}\), \(W_{\text{max}}\), \(C_\text{p}\), \(C_{\text{total}}\). Then, to avoid weak intensity, which leads to poor adversariality, or excessive intensity, which causes image distortion, the range for the post-processing intensity is set. We formulate the $k$-th particle swarm as \(P^k=(p_1^k,p_2^k,\dots,p_N^k)\). 

Each particle in the particle swarm, such as \(p_i^k\), the $i$-th particle of $k$-th iteration, has four attributes: 

\textit{1) The position \(Pos(p_i^k)\):} This is the post-processing fusion intensity of the particle, which is formulated as Eq. \ref{eq:pos_param}.
\begin{equation}
    \label{eq:pos_param}
    Pos(p_i^k)=(\alpha, \beta, \phi, \sigma, \lambda_{x},\lambda_{y},\theta,\gamma),
\end{equation}
where \(k=1\) and these 8 parameters represent the intensity of the four post-processing, randomly generated by Eq. \ref{eq:generated_param}. In Eq. \ref{eq:generated_param}, \(\tau\in\{\alpha,\beta,\phi,\sigma,\lambda_{x},\lambda_{y},\theta,\gamma\}\), \(\tau_{\text{min}}\) and \(\tau_{\text{max}}\) are the lower and upper bounds of the corresponding parameters.
\begin{equation}
    \label{eq:generated_param}
    \tau=random(\min=\tau_{\text{min}}, \max=\tau_{\text{max}})
\end{equation}

\begin{algorithm}[H]
    \caption{R$^2$BA optimization process.}
    \label{alg:Method}
    \begin{algorithmic}[1]
        \REQUIRE 
            Original fake image $\mathbf{X}$
        \ENSURE 
            Adversarial example $\mathbf{X}_{\text{adv}}$
        \STATE{Initialize global parameters \(N\), \(K\)}
        \STATE{Set \(W_{\text{min}}\), \(W_{\text{max}}\), \(C_\text{p}\), \(C_{\text{total}}\)}
        \STATE{Set the ranges of \(\alpha,\beta,\phi,\sigma,\lambda_{x},\lambda_{y},\theta,\gamma\)}
        \STATE{Initialize particles by Eq.\ref{eq:pos_param} to Eq.\ref{eq:v_init}}
        \FOR{$i = 1, 2, \dots, N$}
            \STATE Calculate fitness \(f_i^1\) of \(p_i^1\) with $\mathbf{X}$ by Eq.\ref{eq:fit_calculte}
            \STATE Set \(f_{i,\text{best}}\) and \(Pos_{\text{best}}(p_i)\) by Eq.\ref{eq:set_fit_init} and Eq.\ref{eq:set_pos_init}
        \ENDFOR 
        \STATE \( f_{\text{best}} = \min_{\substack{i=1,2,\dots,N}} \{f_{i,\text{best}}\}\)
        \STATE{set \(Pos_{\text{best}}(P)\) corresponding to \(f_{\text{best}}\)}
        \STATE{Set current iteration \(k=2\)}
        \WHILE{$k \leq K \text{ and } f_{\text{best}} \geq 0.5$}
        \FOR{$i = 1, 2, \dots, N$}
                \STATE{Update \(w^k\), \(v_i^k\), \(Pos(p_i^k)\)} by Eq. \ref{eq:inertial_decay} to Eq. \ref{eq:special_2}
                \STATE{Calculate fitness \(f_i^k\) of \(p_i^k\) with $\mathbf{X}$ by Eq.\ref{eq:fit_calculte}}
                \IF{\(f_i^k < f_{i,\text{best}}\)}
                    \STATE{\(f_{i,\text{best}}=f_i^k\)}
                    \STATE{\(Pos_{\text{best}}(p_i)=Pos(p_i^k)\)}
                \ENDIF
                \IF{\(f_{i,\text{best}} < f_{\text{best}}\)}
                    \STATE{\(f_{\text{best}}=f_{i,\text{best}}\)}
                    \STATE{\(Pos_{\text{best}}(P)=Pos_{\text{best}}(p_i)\)}
                \ENDIF
            \ENDFOR
        \ENDWHILE
        \STATE Select \(Pos_{\text{adv}}(P)\) by Eq. \ref{eq:selection} and Eq. \ref{eq:ssim}
        \STATE{$\mathbf{X}_{\text{adv}}$= $\mathbf{X}_{Pos_{\text{adv}}(P)}$}
        \RETURN $\mathbf{X}_{\text{adv}}$
    \end{algorithmic}
\end{algorithm}

\textit{2) The velocity \(v_i^k\):} The velocity of each particle will be initialized according to Eq. \ref{eq:v_define} and Eq. \ref{eq:v_init}.
\begin{equation}
\label{eq:v_define}
v_i^k=(v_{i(1)}^k, v_{i(2)}^k, \dots, v_{i(M)}^k), M=8
\end{equation}
\begin{equation}
\label{eq:v_init}
v_{i(j)}^1=random(\min=-1, \max=1)
\end{equation}
where \(v_i^k\) is the velocity of $i$-th particle of the $k$-th iteration, and \(v_{i(j)}^k\) is the $j$-th latitude velocity of \(v_i^k\). \(M\) is the dimension of the problem. Since we have 8 parameters, \(M=8\). 

\textit{3) The fitness \(f_i^k\):} We calculate the fitness of each particle by Eq. \ref{eq:fit_calculte}. \(X_{Pos(p_i^k)}\) denotes the post-processed image of \(X\), and \(Pos(p_i^k)\) is the post-processing fusion intensity. 
\begin{equation}
\label{eq:fit_calculte}
f_i^k=D(X_{Pos(p_i^k)})
\end{equation}
%where \(D(\cdot)\) is the victim detector with a fake probability output. 
In other words, a smaller fitness value indicates a lower fake probability, implying better adversariality.

\textit{4) The individual best \(f_{i,\text{best}}\) and \(Pos_{\text{best}}(p_i)\):} The individual best refers to the best position 
\(Pos_{\text{best}}(p_i)\) corresponding to its all-time best fitness \(f_{i,\text{best}}\). In initialization, we set the individual best by Eq. \ref{eq:set_fit_init} and Eq. \ref{eq:set_pos_init} and define the global best \(f_{\text{best}}=\min\left\{f_{i,\text{best}}\right\}\) and 
\(Pos_{\text{best}}(P)\) as the particle position corresponding to \(f_{\text{best}}\). 
\begin{equation}
\label{eq:set_fit_init}
f_{i,\text{best}}=f_i^1, i=1,2,\dots,N
\end{equation}
\begin{equation}
\label{eq:set_pos_init}
Pos_{\text{best}}(p_i)=Pos(p_i^1), i=1,2,\dots,N
\end{equation}

{\textbf{Optimization iteration process.}} 
Now we have completed the initialization. Here comes the optimization iteration. Before each iteration begins, we need to determine particle's current position \(Pos(p_i^k)\) and its next velocity \(V(p_i^{k+1})\), which are the key factors that influence the particle's next position \(Pos(p_i^{k+1})\). Here \(Pos(p_i^k)\) is known. Therefore, the focus is on how to update the velocity of the particles. The update of the particle's velocity involves two aspects: inertia \(w\) and the two best \(Pos_{\text{best}}(p_i)\) and \(Pos_{\text{best}}(P)\). The following describes how to update the particle's velocity and position respectively. 

\textit{1) Velocity update} 

The inertia weight \(w\) controls how much a particle depends on its previous velocity during updates. By gradually decreasing the inertia weight, R$^2$BA can take different convergence modes during the optimization: Early in the optimization, larger inertia weight accelerates global exploration, enhances the ability to explore the entire example domain space, and reduces the risk of local optima. As optimization progresses and inertia decreases, particles can more precisely converge on the global best, reducing unnecessary position shifts and improving both convergence speed and accuracy. In R$^2$BA, using linear decay allows the particles to gradually and smoothly move towards the global best solution. The formula for inertia decay is given in Eq. \ref{eq:inertial_decay}.
\begin{equation}
\label{eq:inertial_decay}
w^k = W_{\text{max}} - (W_{\text{max}}-W_{\text{min}})*\frac{k}{K}
\end{equation}

The particle's velocity controls its movement. We want the particle to move towards the decision boundary (where the fake probability is 0.5) and even across it to execute a successful attack. Therefore, it is the most direct and effective method to guide the velocity update with the individual best and the global best, which are always updated to the position with the smallest fake probability. Therefore, we use the distance between \(Pos(p_i^k)\) and \(Pos_{\text{best}}(p_i)\)/\(Pos_{\text{best}}(P)\) as guidance to update the particle's velocity, which in turn influences the particle's position in the next iteration. Combined with the inertia \(w\), the velocity is update by Eq. \ref{eq:velocity_update}, Eq. \ref{eq:velocity_p} and Eq. \ref{eq:velocity_total}. 
\begin{equation}
\label{eq:velocity_update}
v_i^{k+1}=w^k\cdot v_i^{k}+C_\text{p}\cdot r_\text{p} \cdot V_\text{p}+C_\text{{total}}\cdot r_\text{{total}} \cdot V_\text{{total}}
\end{equation}
\begin{equation}
\label{eq:velocity_p}
V_\text{p}=Pos_{\text{best}}(p_i)-Pos(p_i^{k})
\end{equation}
\begin{equation}
\label{eq:velocity_total}
V_\text{{total}}=Pos_{\text{best}}(P)-Pos(p_i^{k})
\end{equation}
where \(r_\text{p}\) and \(r_\text{total}\) are random decimals in \([0,1)\), and \(V_\text{p}\) and \(V_\text{{total}}\) represent the influence of the individual best position and the global best position from $k$-th iteration on the particle's velocity in $k+1$-th iteration, respectively. 

\textit{2) Position update} 

Next, we update the particle's position based on the updated velocity \(v_i^{k+1}\), the current position \(Pos(p_i^{k})\) and a random modification term \(\varepsilon_i\) by Eq. \ref{eq:position_update} and Eq. \ref{eq:modification}. The particle velocity incorporates both individual best and global best experiences, which are the key factors driving the particle towards the decision boundary.
\begin{equation}
\label{eq:position_update}
Pos(p_i^{k+1})=Pos(p_i^{k})+v_i^{k+1}+\varepsilon_i
\end{equation}
\begin{equation}
\label{eq:modification}
\varepsilon_i =
\begin{cases}
random(),& \text{with prob } \epsilon, \\
0,& \text{with prob } 1-\epsilon
\end{cases}
\end{equation}
Here \(\varepsilon_i\) is updated every iteration to increase the diversity of particles. And we set probability \(\epsilon\) to 0.5, and \(random()\) can output a decimal in the range \([0, 1)\) for modification.

Since the updated position \(Pos(p_i^{k+1})\) may exceed the defined parameter range, we need to set constraints on the particles' positions. The basic constraints are given in Eq. \ref{eq:constraints}. For certain parameters, such as the radius and coordinates, we need to ensure that they are integers or odd numbers. These constraints are described in Eq. \ref{eq:special_1} and Eq. \ref{eq:special_2}.
\begin{equation}
\label{eq:constraints}
\tau=\min(\max(\tau,\tau_{\text{min}}), \tau_{\text{max}})
\end{equation}
\begin{equation}
\label{eq:special_1}
\tau=Integer(\tau), \tau \in \{\alpha,\phi,\lambda_{x},\lambda_{y},\gamma\}
\end{equation}
\begin{equation}
\label{eq:special_2}
\tau=Odd(\tau), \tau \in \{\alpha\}
\end{equation}

The updates have now been completed. And next we follow the steps outlined in lines 15 to 23 of Algorithm \ref{alg:Method} to optimize the individual best of particles and the global best of the particle swarm. Our optimization goal is to minimize the fake probability of all adversarial particles, as shown in Eq. \ref{eq:min_fitness}.
\begin{equation}
  \label{eq:min_fitness}
  \min(f_\text{best}) \Rightarrow \underset{\text{N}}{\min} \left\{f_{i,\text{best}}\right\} \Rightarrow \underset{\text{N}}{\min} D(X_{Pos_{\text{best}}(p_i)})
\end{equation}
Specifically, we use Eq. \ref{eq:fit_calculte} to calculate the fake probability as the particle's fitness, and compare them one by one to determine the individual best and the global best. 

If the global best fitness is greater than 0.5 and the iteration has not reached the max iteration, the algorithm proceeds to the next \textit{optimization iteration}. Otherwise, go to \textit{selection}. 

{\textbf{Selection.}} We use Eq. \ref{eq:selection} to determine the final output \(Pos_{\text{adv}}(P)\). When the global best fitness is larger than \(0.5\), we let \(Pos_{\text{adv}}(P)\) to be the particle with the smallest fitness. Otherwise, we select the particle with the largest SSIM\cite{SSIM} from the particles with successful attacks. The \(ssim(\cdot)\)\cite{SSIM} calculate the structural similarity between images, where the larger the value, the better the image quality. 
\begin{equation}
\label{eq:selection}
Pos_{\text{adv}}(P) =
\begin{cases}
Pos_{\text{best}}(P),& f_  {\text{best}} \geq 0.5, \\
Pos_{\text{ssim}},& otherwise
\end{cases}
\end{equation}
\begin{equation}
\label{eq:ssim}
Pos_{\text{ssim}} = Pos_{max\{ssim(\mathbf{X}_{Pos_{\text{best}}(p_i)}, \mathbf{X})|f_{i,\text{best}} < 0.5\}}
\end{equation}

\section{Experiments}\label{section:Experiments}
\begin{table}
    \renewcommand\arraystretch{1.0}
    \caption{GAN-based and Diffusion-based AIGC Datasets\label{tab:dataset}}
    \begin{center}
    \begin{tabular}{p{1.8cm}<{\centering} p{1.3cm}<{\centering} p{1.8cm}<{\centering} p{1.8cm}<{\centering}}
    \toprule  
    \textbf{Real Source}& \textbf{Fake Type}& \textbf{Models}& \textbf{Data source} \\
    \midrule
    ImageNet\cite{ImageNet}& GAN& ProGAN\cite{ProGAN}& CNNSpot\cite{CNNSpot} \\
    ImageNet\cite{ImageNet}& Diffusion& VQDM\cite{VQDM}& GenImage\cite{GenImage} \\
    MSCOCO\cite{MSCOCO}& Diffusion& LDM\cite{LDM}& DRCT\cite{DRCT} \\
    \bottomrule
    \end{tabular}
    \end{center}
\end{table}

\subsection{Experimental Setups}
{\textbf{Datasets.}} We conduct experiments on 3 datasets to evaluate our method. For GAN-based forgery detection, we use CNNSpot\cite{CNNSpot} as the GAN-based AIGC dataset. For diffusion-based forgery detection, we use GenImage\cite{GenImage} and DRCT\cite{DRCT} as the diffusion-based AIGC datasets. All 3 datasets are generated using publicly available forgery models. Table \ref{tab:dataset} shows the statistics of the AIGC datasets we used. 

{\textbf{Victim detectors.}} We use CNNSpot\cite{CNNSpot}, ResNetND\cite{ResNetND}, UnivFD\cite{UnivFD}, and FatFormer\cite{FatFormer} for GAN-based detection, and FatFormer\cite{FatFormer}, UnivFD pretrained by DRCT\cite{DRCT}, and ConvB\cite{ConvB} pretrained by DRCT\cite{DRCT} for diffusion-based detection. Here the FatFormer keeps the same on all datasets. Though UnivFD and UnivFD(DRCT) share the same model architecture, the specific models employed are different. We show these detectors' accuracy in Table \ref{tab:ori_acc_GAN} and Table \ref{tab:ori_acc_Diffusion}.

\begin{table}
    \setlength{\tabcolsep}{0mm}{
    \renewcommand\arraystretch{1.0}
    \caption{The accuracy of GAN-based AIGC detectors \label{tab:ori_acc_GAN}}
    \begin{center}
    \begin{tabular}{ p{1.7cm}<{\centering} p{1.75cm}<{\centering} p{1.8cm}<{\centering} p{1.65cm}<{\centering} p{1.8cm}<{\centering}}
    \toprule  
    \multirow{2}{*}{\textbf{Datasets}}& \multicolumn{4}{c}{\textbf{Detectors}} \\
    \cmidrule(l){2-5}
  &CNNSpot\cite{CNNSpot}&ResNetND\cite{ResNetND}&UnivFD\cite{UnivFD}&FatFormer\cite{FatFormer}\\
    \midrule
    CNNSpot\cite{CNNSpot}& 0.993& 0.947& 0.997& 0.998 \\
    \bottomrule
    \end{tabular}
    \end{center}}
\end{table}
\begin{table}
    \setlength{\tabcolsep}{0mm}{
    \renewcommand\arraystretch{1.0}
    \caption{The accuracy of Diffusion-based AIGC detectors \label{tab:ori_acc_Diffusion}}
    \begin{center}
    \begin{tabular}{p{1.9cm}<{\centering}  p{1.9cm}<{\centering} p{2.5cm}<{\centering} p{2.4cm}<{\centering}}
    \toprule  
     \multirow{2}{*}{\textbf{Datasets}}& \multicolumn{3}{c}{\textbf{Detectors}} \\
    \cmidrule(l){2-4}
  & FatFormer\cite{FatFormer}& UnivFD(DRCT)\cite{DRCT}& ConvB(DRCT)\cite{DRCT} \\
    \midrule
     GenImage\cite{GenImage}& 0.810& 0.891& 0.769 \\
     DRCT\cite{DRCT}& 0.963& 0.967& 0.999 \\
    \bottomrule
    \end{tabular}
    \end{center}}
\end{table}

\begin{table*}
    \setlength{\tabcolsep}{0.35mm}{
    \renewcommand\arraystretch{1.1}
    \caption{The attack success rate and image quality results on GAN-based AIGC detectors \label{tab:nondiff}}
    \begin{center}
    \begin{tabular}{p{0.9cm}<{\centering} p{1.9cm}<{\centering} 
    p{0.93cm}<{\centering} p{0.93cm}<{\centering} p{0.93cm}<{\centering}
    p{0.93cm}<{\centering} p{0.93cm}<{\centering} p{0.93cm}<{\centering}
    p{0.93cm}<{\centering} p{0.93cm}<{\centering} p{0.93cm}<{\centering}
    p{0.93cm}<{\centering} p{0.93cm}<{\centering} p{0.93cm}<{\centering}
    p{0.93cm}<{\centering} p{0.93cm}<{\centering} p{0.93cm}<{\centering}
    }
    \toprule  
     \multirow{2}{*}{\textbf{Dataset}}& \multirow{2}{*}{\textbf{Attack}} 
  & \multicolumn{3}{c}{\textbf{CNNspot}\cite{CNNSpot}} 
  & \multicolumn{3}{c}{\textbf{ResNetND}\cite{ResNetND}}  
  & \multicolumn{3}{c}{\textbf{UnivFD}\cite{UnivFD}}  
  & \multicolumn{3}{c}{\textbf{FatFormer}\cite{FatFormer}}  
  & \multicolumn{3}{c}{\textbf{Average}}  
     \\
    \cmidrule(lr){3-5}
    \cmidrule(lr){6-8}
    \cmidrule(lr){9-11}
    \cmidrule(lr){12-14}
    \cmidrule(l){15-17}
  &&
    ASR↑&BRIS↓&SSIM↑&
    ASR↑&BRIS↓&SSIM↑&
    ASR↑&BRIS↓&SSIM↑&
    ASR↑&BRIS↓&SSIM↑&
    ASR↑&BRIS↓&SSIM↑\\
    \cmidrule(){1-2}\cmidrule(lr){3-5}\cmidrule(lr){6-8}\cmidrule(lr){9-11}\cmidrule(lr){12-14}\cmidrule(l){15-17}
    \multirow{8}{*}{\rotatebox{90}{\textbf{CNNSpot}\cite{CNNSpot}}} 
  & FGSM\cite{FGSM}& 0.565& 42.367& 0.687& 0.590& 42.453& \underline{0.688}&0.015&\underline{40.071}&\textbf{0.704}& 0.131& 40.364& \underline{0.703}& 0.325& 41.314& \underline{0.696}\\
  & MIFGSM\cite{MIFGSM}& 0.575& 42.137& 0.688& 0.570& 42.596& \underline{0.688}& 0.020& 40.624& \underline{0.703}& 0.090& \underline{40.339}& \underline{0.703}& 0.314& 41.424& \underline{0.696}\\
  & PGD\cite{PGD}& 0.580& 42.181& 0.688& 0.580& 42.298& 0.687& 0.010& 40.543& \underline{0.703}& 0.116& 40.567& 0.702& 0.321& 41.397& 0.695\\
  & StatAT\cite{Evade}& 0.340& \underline{33.521}& 0.634& 0.390& \underline{28.235}& 0.629& 0.165& \textbf{39.148}& 0.602& 0.146& 45.288& 0.535& 0.260& \underline{36.548}& 0.600\\
    \cmidrule(l){2-2}\cmidrule(lr){3-5}\cmidrule(lr){6-8}\cmidrule(lr){9-11}\cmidrule(lr){12-14}\cmidrule(l){15-17}
  & SimBA\cite{SimBA}& 0.475& 49.347& \underline{0.691}& \underline{0.830}& 50.700& 0.634& 0.010& 50.031& 0.574& 0.161& 65.631& 0.457& 0.369& 53.927& 0.589\\
  & Square\cite{Square}& \textbf{1.000}& 112.784& 0.534& \textbf{1.000}& 115.384& 0.533& \underline{0.605}& 67.305& 0.617& \underline{0.623}& 79.314& 0.584& \underline{0.807}& 93.697& 0.567\\
  & BruSLe\cite{BruSLe}& \underline{0.925}&76.541&0.638&	0.820 &	76.692&	0.638&	0.605&71.369&0.584&	0.347&71.243&0.583&	0.674&73.961&0.611 \\
    \cmidrule(l){2-2}\cmidrule(lr){3-5}\cmidrule(lr){6-8}\cmidrule(lr){9-11}\cmidrule(lr){12-14}\cmidrule(l){15-17}
  & \textbf{R$^2$BA}& \textbf{1.000}& \textbf{27.352}& \textbf{0.699}& \textbf{1.000}& \textbf{21.066}& \textbf{0.745}& \textbf{0.620}& 41.380& 0.642& \textbf{1.000}& \textbf{38.681}& \textbf{0.778}& \textbf{0.905}& \textbf{32.120}& \textbf{0.716} \\
    \bottomrule
    \end{tabular}
    \end{center}}
\end{table*}
\begin{table*}
    \setlength{\tabcolsep}{1mm}{
    \renewcommand\arraystretch{1.0}
    \caption{The attack success rate and image quality result on Diffusion-based AIGC detectors \label{tab:diff}}
    \begin{center}
    \begin{tabular}{
    p{1cm}<{\centering} p{2cm}<{\centering} 
    p{1cm}<{\centering} p{1cm}<{\centering} p{1cm}<{\centering}
    p{1cm}<{\centering} p{1cm}<{\centering} p{1cm}<{\centering}
    p{1cm}<{\centering} p{1cm}<{\centering} p{1cm}<{\centering}
    p{1cm}<{\centering} p{1cm}<{\centering} p{1cm}<{\centering}
    }
    \toprule 
    \multirow{2}{*}{\textbf{Datasets}}&  \multirow{2}{*}{\textbf{Attacks}} 
  & \multicolumn{3}{c}{\textbf{FatFormer}\cite{FatFormer}} 
  & \multicolumn{3}{c}{\textbf{UnivFD(DRCT)}\cite{DRCT}} 
  & \multicolumn{3}{c}{\textbf{ConvB(DRCT)}\cite{DRCT}} 
  & \multicolumn{3}{c}{\textbf{Average}} \\
    \cmidrule(lr){3-5}
    \cmidrule(lr){6-8}
    \cmidrule(lr){9-11}
    \cmidrule(l){12-14}
  &&
    ASR↑&BRIS↓&SSIM↑&
    ASR↑&BRIS↓&SSIM↑&
    ASR↑&BRIS↓&SSIM↑&
    ASR↑&BRIS↓&SSIM↑\\
    \cmidrule(lr){1-2}\cmidrule(lr){3-5}\cmidrule(lr){6-8}\cmidrule(lr){9-11}\cmidrule(l){12-14}
    \multirow{8}{*}{\rotatebox{90}{\textbf{GenImage}\cite{GenImage}}}
  & FGSM\cite{FGSM}& 0.406& 53.226& \underline{0.643}& 0.027& 53.280& \underline{0.626}& 0.311& 53.089& 0.622& 0.248& 53.198& \underline{0.630}  \\
  & MIFGSM\cite{MIFGSM}& 0.394& 52.903& 0.640& 0.032& 52.930& \underline{0.626}& 0.340& 53.005& 0.622& 0.255& 52.946& 0.629\\
  & PGD\cite{PGD}& 0.406& 52.722& 0.642& 0.043& 53.008& 0.625& 0.311& \underline{52.890}& \underline{0.623}& 0.254& 52.873& \underline{0.630}  \\
  & StatAT\cite{Evade}& 0.244& \underline{49.824}& 0.509& 0.210& 54.794& 0.552& 0.925& 70.335& 0.531& 0.459& 58.318& 0.531\\
    \cmidrule(lr){2-2}\cmidrule(lr){3-5}\cmidrule(lr){6-8}\cmidrule(lr){9-11}\cmidrule(l){12-14}
  & SimBA\cite{SimBA}& 0.231& 57.114& 0.522& 0.102& \underline{45.556}& 0.334& 0.811& 46.938& 0.497& 0.382& \underline{49.869}& 0.451\\
  & Square\cite{Square}& \underline{0.588}& 83.995& 0.523& \underline{0.823}& 105.001& 0.519& 0.962& 107.737& 0.489& \underline{0.791}& 98.911& 0.510\\
  & BruSLe\cite{BruSLe}&0.500&70.989& 0.536& 0.780&69.553 &0.553&\underline{0.972}& 70.268& 0.548& 0.750&70.270&0.546\\
    \cmidrule(lr){2-2}\cmidrule(lr){3-5}\cmidrule(lr){6-8}\cmidrule(lr){9-11}\cmidrule(l){12-14}
  & \textbf{R$^2$BA}& \textbf{1.000}& \textbf{25.209}& \textbf{0.827}& \textbf{0.898}& \textbf{32.098}& \textbf{0.665}& \textbf{1.000}& \textbf{19.494}& \textbf{0.804}& \textbf{0.966}& \textbf{25.600}& \textbf{0.765} \\
    \midrule
    \multirow{8}{*}{\rotatebox{90}{\textbf{DRCT}\cite{DRCT}}}
  & FGSM\cite{FGSM}& 0.302& 38.376& 0.725& 0.020& 38.511& \textbf{0.726}& 0.005& 38.489& \textbf{0.725}& 0.109& 38.459& \underline{0.725}  \\
  & MIFGSM\cite{MIFGSM}& 0.271& 38.488& \underline{0.726}& 0.025& 38.551& \underline{0.724}& 0.000& 38.746& \underline{0.724}& 0.099& 38.595& \underline{0.725}  \\
  & PGD\cite{PGD}& 0.286& 38.158& 0.725& 0.025& \underline{38.384}& 0.723& 0.000& \underline{38.103}& 0.723& 0.104& \underline{38.215}& 0.724\\
  & StatAT\cite{Evade}& 0.246& \underline{35.319}& 0.547& 0.191& 40.615& 0.609& 0.760& 45.700& 0.592& 0.399& 40.545& 0.582\\
    \cmidrule(lr){2-2}\cmidrule(lr){3-5}\cmidrule(lr){6-8}\cmidrule(lr){9-11}\cmidrule(l){12-14}
  & SimBA\cite{SimBA}& 0.196& 71.740& 0.433& 0.010& 63.628& 0.376& 0.565& 69.004& 0.376& 0.257& 68.124& 0.395\\
  & Square\cite{Square}& \underline{0.618}& 78.449& 0.618& \underline{0.628}& 85.965& 0.620& \underline{0.875}& 92.046& 0.623& \underline{0.707}& 85.487& 0.620\\
  & BruSLe\cite{BruSLe}& 0.523 &64.247 &0.630 &0.568 &61.990 &0.648 &0.490 &62.362 &0.644 &0.527 &62.866 &0.641 \\
    \cmidrule(lr){2-2}\cmidrule(lr){3-5}\cmidrule(lr){6-8}\cmidrule(lr){9-11}\cmidrule(l){12-14}
  & \textbf{R$^2$BA}& \textbf{1.000}& \textbf{28.764}& \textbf{0.823}& \textbf{0.693}& \textbf{37.029}& 0.664& \textbf{0.955}& \textbf{29.609}& 0.699& \textbf{0.883}& \textbf{31.801}& \textbf{0.728} \\
    \bottomrule
    \end{tabular}
    \end{center}}
\end{table*}

{\textbf{Metrics.}} We evaluate the effectiveness of the attacks using the attack success rate (ASR) and image quality metrics, BRISQUE\cite{BRISQUE} and SSIM\cite{SSIM}. ASR measures the percentage of AI-generated images misclassified as natural (fake probability\textless 0.5). BRISQUE is a non-reference image quality metric, with higher scores indicating poorer quality, while SSIM measures structural similarity between images, with values closer to 1 indicating better quality. 

{\textbf{Basic attacks on white box and black box.}} We compare four white-box attacks, including FGSM\cite{FGSM}, MIFGSM\cite{MIFGSM}, PGD\cite{PGD} and StatAT\cite{Evade}, and three black-box attacks, including SimBA\cite{SimBA}, Square\cite{Square} and BruSLe\cite{BruSLe}. StatAT\cite{Evade} is the SOTA white-box post-processing-based attack, and BruSLe\cite{BruSLe} is the SOTA black-box query-based attack. For the parameters of baseline attacks, we set the max perturbation magnitude to \(20/255\) with the pixel range \([0,1]\), the iterations number to \(20\), the max number of queries is \(1000\) and use random initialization to start attacks.

{\textbf{Implementation details.}} About image size, except for attacks on CNNSpot\cite{CNNSpot} and ResNetND\cite{ResNetND} with a \(256\times256\) image size, all other models use center-cropped \(224\times224\) images to match the suitable input size of the pretrained models. For better analyses, we use {\textbf{bold}} font to indicate the best result and {\underline{underline}} the second best result. We set the max number of queries to 1000 and others are set as follows:
\begin{enumerate}
\item{Global parameters: \(N=100\), \(K=10\), \(W_{\text{min}}=1\), 
\(W_{\text{max}}=5\), \(C_\text{p}=C_{\text{total}}=1.5\)}
\item{Gaussian blur: \(\alpha\in\{1,3,\dots,13\}\), \(\beta\in[0.5,5]\)}
\item{JPEG compression: \(\phi\in\{10,\dots,100\}\).}
\item{Gaussian noise: \(\sigma\in[0.0001,(5/255)^2]\).}
\item{Light spot: \(\theta\in[0.2,1.8]\), \(\gamma\in\{30,\dots,100\}\) and 
the spot center \((\lambda_{x},\lambda_{y})\) is controlled in the image.}
\end{enumerate}

\subsection{Anti-detection Performance Analysis \label{anti-perform}}
In this subsection, we show the anti-detection performance of attacking AIGC detectors on three datasets. In Table \ref{tab:nondiff} and Table \ref{tab:diff}, we present ASR, the image quality metrics BRISQUE (abbreviated as BRIS) and SSIM, and compare the average performance of each metric on all detectors.

{\textbf{Attacks on GAN-based detectors.}} The experiments in Table \ref{tab:nondiff} prove that the attack success rate of R$^2$BA reaches the SOTA level, with a max improvement of 38\% in ASR and an average ASR improvement of about 10\%. In terms of image quality, it is at the highest level among the average results and all GAN-based detectors except for the UnivFD\cite{UnivFD} detector, indicating the excellent invisibility of our proposed method. Since we limited the max perturbation magnitude for gradient-based attacks, their image quality on UnivFD is slightly better than ours. However, the attack effectiveness of these white-box methods against the UnivFD detector is nearly zero, confirming that gradient-based white-box attacks struggle to achieve effective anti-detection performance.

{\textbf{Attacks on Diffusion-based detectors.}} The results in Table \ref{tab:diff} show that, on the GenImage\cite{GenImage} and DRCT\cite{DRCT} datasets, R$^2$BA has a maximum improvement of 41\% in ASR and an average ASR improvement of approximately 18\%. Additionally, in terms of the average image quality, R$^2$BA achieves an improvement of 33\%--69\% in BRISQUE\cite{BRISQUE} and 11\%--77\% in SSIM\cite{SSIM}, indicating that R$^2$BA demonstrates excellent invisibility when attacking diffusion-based AIGC detectors.

{\textbf{Comprehensive analysis.}} The results of on three popular AIGC datasets show that perturbations generated by traditional gradient-based or random pixel-modification attacks\cite{BruSLe},\cite{FGSM},\cite{MIFGSM},\cite{PGD},\cite{SimBA} are often detected as fake by well-trained AIGC detectors. As a post-processing-based white-box attack, due to based on statistical feature consistency, StatAT\cite{Evade} performs well on facial images but poorly on multi-class natural images, as this type of methods cannot unify the feature dimensions on natural images with inherent multi-class nature and complexity. While Square\cite{Square} performs better in ASR, it performs worse in image quality metrics compared to other algorithms, suggesting that it may improve anti-detection performance at the expense of image quality.

In contrast, as a post-processing fusion adversarial attack, R$^2$BA does not require consideration of feature-level consistency. Instead, it targets the vulnerabilities of the detectors to achieve efficient attacks, making it an image-class-independent adversarial attack. By optimizing the post-processing fusion intensity, R$^2$BA achieves an average anti-detection performance improvement of 15\%--72\%, along with a significant improvement of 26\%--68\% in the BRISQUE\cite{BRISQUE} across the three datasets. It successfully evades both GAN-based and diffusion-based AIGC detectors. The realistic-like adversarial examples generated by R$^2$BA, which match the pattern of the real-world images, cause the detectors to classify them as real natural images. In other words, when dealing with post-processing fusion attacks, the existing AIGC detectors suffer significant performance degradation, which suggests that if the AIGC detectors are put into practical application, their security cannot be guaranteed and requires further investigation.

\subsection{Robustness Comparison Experiments}
The robustness of adversarial examples to common image post-processing techniques is crucial for their real-world applicability. If an adversarial example is vulnerable to such techniques, its adversariality is erased, and the threat to AIGC detectors is eliminated. 
\begin{table*}
    \setlength{\tabcolsep}{0.65mm}{
    \renewcommand\arraystretch{1.0}
    \caption{Robustness experiments against known post-processing — JPEG compression and Gaussian noise \label{tab:robust}}
    \begin{center}
    \begin{tabular}{
    p{1.1cm}<{\centering} p{2.4cm}<{\centering} 
    p{1cm}<{\centering} p{1cm}<{\centering} 
    p{1cm}<{\centering} p{1cm}<{\centering} 
    p{1cm}<{\centering} p{1cm}<{\centering}
    p{1cm}<{\centering} p{1cm}<{\centering} 
    p{1cm}<{\centering} p{1cm}<{\centering} 
    p{1cm}<{\centering} p{1cm}<{\centering}
    }
    \toprule  
     \multirow{2}{*}{\textbf{Dataset}}& \multirow{2}{*}{\textbf{Attacks}}& \multicolumn{6}{c}{\textbf{ASR↑ - JPEG compression}}& \multicolumn{6}{c}{\textbf{ASR↑ - Gaussian noise}}\\
    \cmidrule(lr){3-8}\cmidrule(l){9-14}
  &&\textbf{50}&\textbf{60}&\textbf{70}&\textbf{80}&\textbf{90}&\textbf{AVG}&\textbf{1/255}&\textbf{2/255}&\textbf{3/255}&\textbf{4/255}&\textbf{5/255}&\textbf{AVG}\\
    \cmidrule(r){1-2}\cmidrule(lr){3-8}\cmidrule(l){9-14}
    \multirow{8}{*}{\rotatebox{90}{\textbf{CNNSpot}\cite{CNNSpot}}}
  & FGSM\cite{FGSM}& 0.142& 0.190& 0.218& 0.268& 0.262& 0.216& 0.325& 0.330& 0.334& 0.345& 0.357& 0.338     \\
  & MIFGSM\cite{MIFGSM}& 0.152& 0.185& 0.207& 0.269& 0.260& 0.214& 0.318& 0.320& 0.336& 0.336& 0.358& 0.334     \\
  & PGD\cite{PGD}& 0.139& 0.187& 0.215& 0.263& 0.256& 0.212& 0.329& 0.328& 0.331& 0.335& 0.350& 0.335     \\
  & StatAT\cite{Evade}& 0.263& 0.322& 0.224& 0.228& 0.219& 0.251& 0.206& 0.217& 0.240& 0.297& 0.326& 0.257     \\
    \cmidrule(lr){2-2}\cmidrule(lr){3-8}\cmidrule(l){9-14}
  &SimBA\cite{SimBA}&0.149&0.213&0.259&0.324&0.361&0.261&0.314&0.255&0.205&0.189&0.186&0.230\\
  & Square\cite{Square}& \underline{0.521}&\underline{0.527}&\underline{0.538}&\underline{0.568}&\underline{0.640}&\underline{0.559}&\underline{0.594}&\underline{0.553}&\underline{0.543}&\underline{0.540}&\underline{0.525}&\underline{0.551}   \\
  &BruSLe\cite{BruSLe}&0.204&0.233&0.279&0.355&0.437&0.301&0.500&0.431&0.394&0.390&0.381&0.419\\
    \cmidrule(lr){2-2}\cmidrule(lr){3-8}\cmidrule(l){9-14}
  & \textbf{R$^2$BA}& \textbf{0.720} & \textbf{0.753} & \textbf{0.801} & \textbf{0.837} & \textbf{0.875} & \textbf{0.797} & \textbf{0.706} & \textbf{0.598} & \textbf{0.575} & \textbf{0.553} & \textbf{0.547} & \textbf{0.596}  \\
    \cmidrule(r){1-2}\cmidrule(lr){3-8}\cmidrule(l){9-14}
    \multirow{8}{*}{\rotatebox{90}{\textbf{GenImage}\cite{GenImage}}}
  & FGSM\cite{FGSM}& 0.237& 0.233& 0.241& 0.321& 0.427& 0.292& 0.242& 0.195& 0.200& 0.167& 0.139& 0.189     \\
  & MIFGSM\cite{MIFGSM}& 0.182& 0.303& 0.218& 0.323& 0.419& 0.289& 0.242& 0.206& 0.180& 0.164& 0.118& 0.182     \\
  & PGD\cite{PGD}& 0.320& 0.290& 0.282& 0.276& 0.444& 0.322& 0.253& 0.228& 0.181& 0.171& 0.154& 0.197     \\
  & StatAT\cite{Evade}& \underline{0.554}&\underline{0.585}&0.366& 0.375& 0.476& 0.471& 0.394& 0.392& 0.371& 0.375& 0.382& 0.383     \\
    \cmidrule(lr){2-2}\cmidrule(lr){3-8}\cmidrule(l){9-14}
  & SimBA\cite{SimBA}&0.454& 0.323& 0.350& 0.293& 0.372& 0.359& 0.295& 0.249& 0.244& 0.205& 0.208& 0.240     \\
  &Square\cite{Square}&0.493& 0.478& 0.387& 0.420& \underline{0.542}&\underline{0.464}&\underline{0.605}&\underline{0.555}&\underline{0.519}&\underline{0.459}&\textbf{0.436} & \underline{0.515}   \\
  & BruSLe\cite{BruSLe}&0.385& 0.480& \underline{0.408}&\underline{0.536}&0.485& 0.459& 0.592& 0.498& 0.437& 0.346& 0.284& 0.431     \\
    \cmidrule(lr){2-2}\cmidrule(lr){3-8}\cmidrule(l){9-14}
  & \textbf{R$^2$BA}& \textbf{0.654} & \textbf{0.691} & \textbf{0.699} & \textbf{0.748} & \textbf{0.793} & \textbf{0.717} & \textbf{0.712} & \textbf{0.577} & \textbf{0.536} & \textbf{0.474} & \underline{0.402}&\textbf{0.540}  \\
    \cmidrule(r){1-2}\cmidrule(lr){3-8}\cmidrule(l){9-14}
    \multirow{8}{*}{\rotatebox{90}{\textbf{DRCT}\cite{DRCT}}}
  & FGSM\cite{FGSM}&0.112& 0.150& 0.117& 0.188& 0.188& 0.151& 0.104& 0.103& 0.094& 0.075& 0.075& 0.090     \\
  & MIFGSM\cite{MIFGSM}&0.139& 0.142& 0.104& 0.155& 0.191& 0.146& 0.099& 0.096& 0.087& 0.093& 0.075& 0.090     \\
  & PGD\cite{PGD}&0.090& 0.125& 0.115& 0.177& 0.181& 0.138& 0.099& 0.096& 0.078& 0.081& 0.074& 0.085     \\
  & StatAT\cite{Evade}& \underline{0.374}&\underline{0.386}&0.327& \underline{0.400}&\underline{0.455}&\underline{0.388}&0.344& 0.328& 0.320& 0.323& 0.297& 0.322     \\
    \cmidrule(lr){2-2}\cmidrule(lr){3-8}\cmidrule(l){9-14}
  & SimBA\cite{SimBA}&0.353& 0.352& 0.302& 0.177& 0.338& 0.304& 0.204& 0.191& 0.198& 0.188& 0.177& 0.192     \\
  & Square\cite{Square}& \underline{0.374}&0.344& \underline{0.337}&0.177& 0.423& 0.331& \underline{0.447}&\underline{0.385}&\underline{0.349}&\underline{0.331}&\underline{0.326}&\underline{0.368}   \\
  & BruSLe\cite{BruSLe}&0.201& 0.219& 0.189& 0.261& 0.261& 0.226& 0.337& 0.256& 0.210& 0.177& 0.175& 0.231     \\ 
    \cmidrule(lr){2-2}\cmidrule(lr){3-8}\cmidrule(l){9-14}
  & \textbf{R$^2$BA}& \textbf{0.578} & \textbf{0.594} & \textbf{0.628} & \textbf{0.697} & \textbf{0.740} & \textbf{0.647} & \textbf{0.661} & \textbf{0.555} & \textbf{0.508} & \textbf{0.478} & \textbf{0.432} & \textbf{0.527} \\
    \bottomrule
    \end{tabular}
    \end{center}}
\end{table*}
\begin{table*}
    \setlength{\tabcolsep}{0.65mm}{
    \renewcommand\arraystretch{1.0}
    \caption{Robustness experiments against unknown post-processing — Rotation and Resize \label{tab:robust_unseen}}
    \begin{center}
    \begin{tabular}{
    p{1.1cm}<{\centering} p{2.4cm}<{\centering} 
    p{1cm}<{\centering} p{1cm}<{\centering} 
    p{1cm}<{\centering} p{1cm}<{\centering} 
    p{1cm}<{\centering} p{1cm}<{\centering}
    p{1cm}<{\centering} p{1cm}<{\centering} 
    p{1cm}<{\centering} p{1cm}<{\centering} 
    p{1cm}<{\centering} p{1cm}<{\centering}
    }
    \toprule  
     \multirow{2}{*}{\textbf{Dataset}}& \multirow{2}{*}{\textbf{Attacks}}& \multicolumn{6}{c}{\textbf{ASR↑ - Rotation}}& \multicolumn{6}{c}{\textbf{ASR↑ - Resize}}\\
    \cmidrule(lr){3-8}\cmidrule(l){9-14}
  &&\textbf{2}&\textbf{4}&\textbf{6}&\textbf{8}&\textbf{10}&\textbf{AVG}&\textbf{1/2}&\textbf{3/4}&\textbf{5/4}&\textbf{3/2}&\textbf{7/4}&\textbf{AVG}\\
    \cmidrule(r){1-2}\cmidrule(lr){3-8}\cmidrule(l){9-14}
    \multirow{8}{*}{\rotatebox{90}{\textbf{CNNSpot}\cite{CNNSpot}}}
  & FGSM\cite{FGSM}&0.356&0.386&0.396&0.406&0.432&0.395&0.145&0.247&0.248&0.248&0.248&0.227\\
  & MIFGSM\cite{MIFGSM}&0.365&0.389&0.407&0.415&0.430&0.401&0.149&0.240&0.254&0.244&0.250&0.227\\
  & PGD\cite{PGD}&0.362&0.387&0.395&0.412&0.429&0.397&0.147&0.243&0.256&0.255&0.256&0.231\\
  & StatAT\cite{Evade}& 0.257&0.279&0.315&0.362&0.396&0.322&0.248&0.232&0.225&0.225&0.230&0.232\\
    \cmidrule(lr){2-2}\cmidrule(lr){3-8}\cmidrule(l){9-14}
  & SimBA\cite{SimBA}& 0.325&0.336&0.345&0.353&0.366&0.345&0.021&0.080&0.133&0.131&0.136&0.100\\
  & Square\cite{Square}& \underline{0.494}& \underline{0.494}& \underline{0.500}& \underline{0.500}& \underline{0.503}& \underline{0.498}& \underline{0.537}& \underline{0.550}& \underline{0.582}& \underline{0.576}& \underline{0.587}& \underline{0.566}  \\
  & BruSLe\cite{BruSLe}& 0.331&0.346&0.372&0.386&0.396&0.366&0.104&0.250&0.338&0.343&0.345&0.276\\
    \cmidrule(lr){2-2}\cmidrule(lr){3-8}\cmidrule(l){9-14}
  & \textbf{R$^2$BA}& \textbf{0.761} & \textbf{0.719} & \textbf{0.705} & \textbf{0.690} & \textbf{0.533} & \textbf{0.681} & \textbf{0.771} & \textbf{0.827} & \textbf{0.844} & \textbf{0.833} & \textbf{0.832} & \textbf{0.821} \\
    \cmidrule(r){1-2}\cmidrule(lr){3-8}\cmidrule(l){9-14}
    \multirow{8}{*}{\rotatebox{90}{\textbf{GenImage}\cite{GenImage}}}
  & FGSM\cite{FGSM}&0.008&0.022&0.049&0.052&0.042&0.035&0.178&0.231&0.258&0.278&0.265&0.242\\
  & MIFGSM\cite{MIFGSM}&0.012&0.027&0.037&0.045&0.051&0.034&0.165&0.228&0.247&0.254&0.256&0.230\\
  & PGD\cite{PGD}& 0.013&0.022&0.035&0.059&0.049&0.036&0.159&0.228&0.235&0.256&0.241&0.224\\
  & StatAT\cite{Evade}&0.218&0.225&0.210&0.215&0.159&0.206&0.118&0.276&0.233&0.220&0.219&0.213\\
    \cmidrule(lr){2-2}\cmidrule(lr){3-8}\cmidrule(l){9-14}
  & SimBA\cite{SimBA}& 0.120&0.168&0.200&0.210&0.220&0.184&0.089&0.146&0.107&0.104&0.100&0.109\\
  &Square\cite{Square}&\underline{0.245}&\underline{0.302}&\underline{0.310}&\underline{0.322}&\underline{0.335}&\underline{0.303}&\underline{0.310}&\underline{0.319}&\underline{0.308}&\underline{0.307}&\underline{0.308}&\underline{0.310}  \\
  & BruSLe\cite{BruSLe}&0.015&0.015&0.022&0.027&0.030&0.022&0.096&0.182&0.103&0.108&0.096&0.117\\
    \cmidrule(lr){2-2}\cmidrule(lr){3-8}\cmidrule(l){9-14}
  & \textbf{R$^2$BA}& \textbf{0.544} & \textbf{0.518} & \textbf{0.513} & \textbf{0.500} & \textbf{0.428} & \textbf{0.500} & \textbf{0.439} & \textbf{0.613} & \textbf{0.684} & \textbf{0.679} & \textbf{0.678} & \textbf{0.619} \\
    \cmidrule(r){1-2}\cmidrule(lr){3-8}\cmidrule(l){9-14}
    \multirow{8}{*}{\rotatebox{90}{\textbf{DRCT}\cite{DRCT}}}
  &FGSM\cite{FGSM}&0.126&0.190&0.191&0.179&0.194&0.176&0.265&0.379&0.258&0.278&0.265&0.289\\
  &MIFGSM\cite{MIFGSM}&0.122&0.188&0.208&0.205&0.176&0.180&0.264&0.366&0.247&0.254&0.256&0.277\\
  &PGD\cite{PGD}&0.126&0.182&0.191&0.166&0.172&0.167&0.280&0.358&0.235&0.256&0.241&0.274\\
  & StatAT\cite{Evade}& 0.332&0.298&0.297&0.303&0.286&0.303&0.239&0.368&0.338&0.329&0.343&0.323\\
    \cmidrule(lr){2-2}\cmidrule(lr){3-8}\cmidrule(l){9-14}
  & SimBA\cite{SimBA}& 0.173&0.170&0.194&0.222&0.226&0.197&0.158&0.233&0.193&0.234&0.208&0.205\\
  &Square\cite{Square}&\underline{0.410}&\underline{0.413}&\underline{0.410}&\underline{0.432}&\underline{0.423}&\underline{0.417}&\underline{0.460}&\underline{0.497}&\underline{0.493}&\underline{0.521}&\underline{0.499}&\underline{0.494}  \\
  & BruSLe\cite{BruSLe}& 0.181&0.168&0.192&0.191&0.166&0.180&0.263&0.367&0.285&0.291&0.281&0.297\\
    \cmidrule(lr){2-2}\cmidrule(lr){3-8}\cmidrule(l){9-14}
  & \textbf{R$^2$BA}& \textbf{0.622} & \textbf{0.614} & \textbf{0.551} & \textbf{0.531} & \textbf{0.435} & \textbf{0.551} & \textbf{0.495} & \textbf{0.703} & \textbf{0.753} & \textbf{0.753} & \textbf{0.750} & \textbf{0.691} \\
    \bottomrule
    \end{tabular}
    \end{center}}
\end{table*}

Therefore, we further conduct robustness comparison experiments for these adversarial attacks. Since our method is based on four post-processing methods, it is naturally robust to common image post-processing. Table \ref{tab:robust} shows robustness experiments against known post-processing techniques, including JPEG compression and Gaussian noise, while Table \ref{tab:robust_unseen} compares robustness against unknown post-processing techniques, including Rotation and Resize (not part of the four post-processing we select), offering a more comprehensive analysis. AVG in both tables refers to average performance.

{\textbf{Against known image post-processing.}} Apparently, the experimental results in Table \ref{tab:robust} show that R$^2$BA far outperforms other adversarial attacks in most settings for robustness. The results indicate that the adversarial examples generated by R$^2$BA can better maintain their adversariality in practical applications and have stronger adaptability to real-world scenarios. Additionally, the results suggest the effectiveness of our post-processing fusion optimization method. Due to the post-processing fusion optimization used in the attack, the image with the best fusion intensity generated by R$^2$BA, when subjected to another similar post-processing, will deviate from the best post-processing intensity, which theoretically leads to some degree of performance degradation in terms of anti-detection. This is indeed the case. The robustness experiments in Table \ref{tab:robust} further confirm this, providing additional evidence for the effectiveness of the optimization in R$^2$BA. 

{\textbf{Against unknown image post-processing.}} In Table \ref{tab:robust_unseen}, it can be seen that our method remains remarkably robust when dealing with unknown post-processing. This is because, essentially, our proposed method, R$^2$BA, is not a simple additive noise perturbation and will not have its adversariality erased by simple image post-processing techniques, such as rotation or resizing. In contrast, apart from causing severe image distortion\cite{Square}, adversarial attack methods based on simple additive noise perturbations are easily faded or eliminated after image post-processing, leading to attack failure.

Overall, after image processing, although the anti-detection performance is somewhat degraded, the ASR metric of R$^2$BA still shows a 21\%--47\% improvement compared to existing adversarial attack methods. Moreover, the ASR of R$^2$BA in robust scenarios remains around 0.5 or even higher, suggesting that the reliability of the detector remains questionable. On the other hand, the robustness experiments demonstrate that stronger post-processing fusion intensity is not always better. This is consistent with our discovery from the pre-experiments in Fig. \ref{fig:proof} of subsection \ref{subsection:Motivation}. Adversarial examples generated with excessively strong post-processing intensity are heavily distorted, which in turn causes the detector to classify the post-processed AIGC image as fake. This also proves that, to some extent, the robustness of AIGC detectors to widely-used image post-processing techniques is uncertain and fluctuates, indicating a potential security risk. 

\subsection{Visual Comparison and Analysis}
\begin{figure*}[!t]
    \centering
    \includegraphics[width=7in]{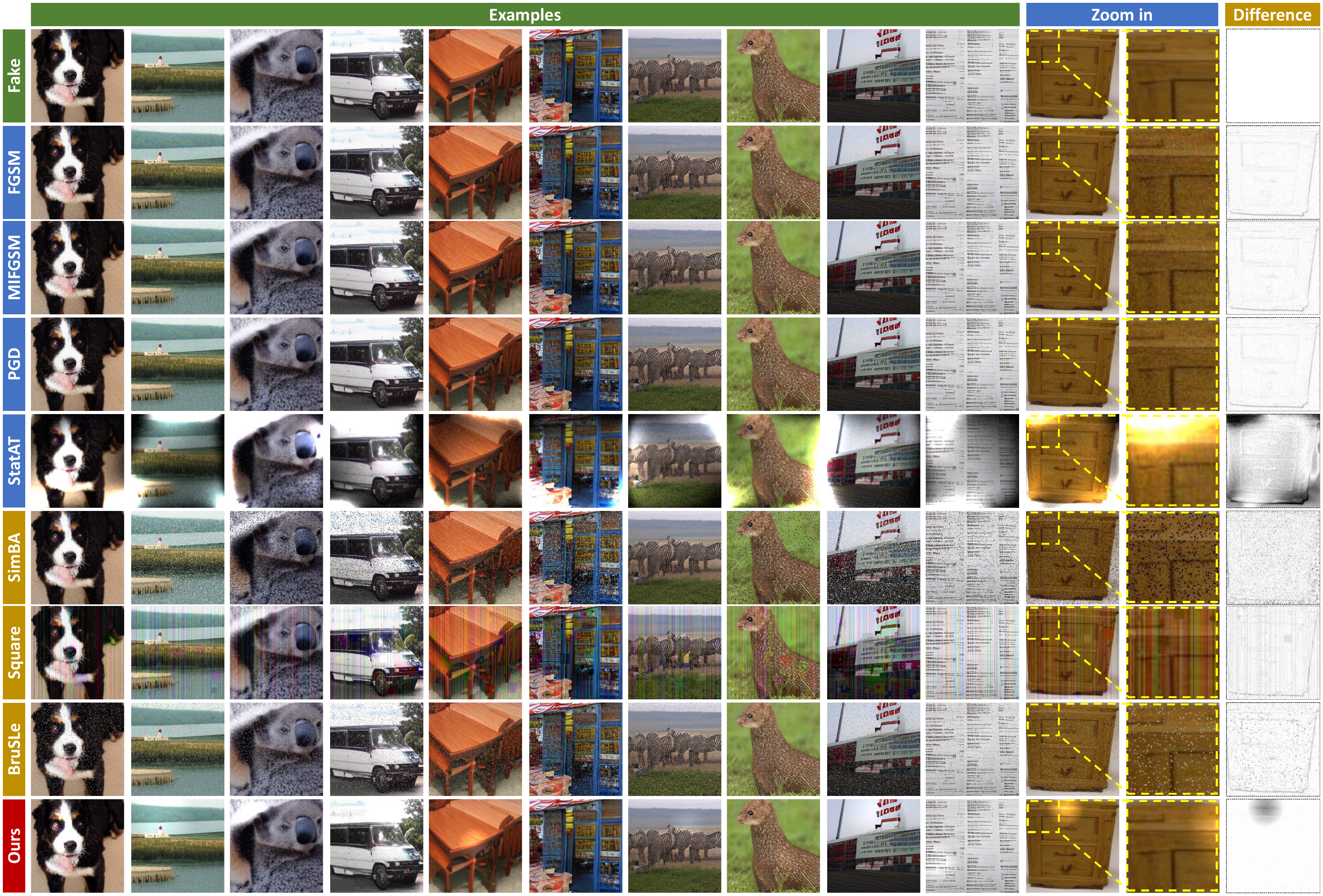}
    \caption{Comparative experiments on image quality of adversarial examples. The dataset we use is GenImage\cite{GenImage} and the detector is FatFormer\cite{FatFormer}. We zoom in on the local details at the end and visualize the perturbations across the entire image.}
    \label{fig:visualize}
\end{figure*}
The invisibility of the adversarial example is crucial in determining whether this perturbation is likely to be detected. Better visualization increases the likelihood of this adversarial example being shared in real life or on the internet. Therefore, we provide a visual comparison in Fig. \ref{fig:visualize}. Ten images from different classes (e.g., dogs, landscapes, cars, letters, etc.) are randomly chosen for the image quality comparison, and one additional image is selected for detailed comparison and perturbation visualization. Combined with the excellent anti-detection performance of R$^2$BA analyzed in subsection \ref{anti-perform}, we can observe that R$^2$BA also maintains high image quality, demonstrates strong image realism, and achieves multi-class anti-detection performance on natural images. 

It is worth noting that, to ensure objectivity and fairness, some images with obvious post-processing traces (e.g., visible light spot) are specially selected for comparison. From both the detailed comparison and perturbation visualization, our image quality is clearly superior. Moreover, the post-processed AIGC images have practical significance. For example, our light spot can simulate a real-world flashlight, while JPEG compression and Gaussian blur mimic real-world image quality degradation. These modifications enhance the realism of adversarial examples, whether viewed from a human or detector's perspective. Additionally, the visualization also shows that existing AIGC detectors are not robust in detecting post-processed AIGC images, and slight image post-processing operations may cause the detector to output incorrect results.

\subsection{Black-box Query Count Comparison Experiments}
\begin{table}
    \setlength{\tabcolsep}{0.35mm}{
    \renewcommand\arraystretch{1.0}
    \caption{Black box Average queries on FatFormer \label{tab:query}}
    \begin{center}
    \begin{tabular}{p{1.75cm}<{\centering} p{1.53cm}<{\centering} p{1.9cm}<{\centering} p{1.9cm}<{\centering} p{1.35cm}<{\centering}}
    \toprule  
    \textbf{Detector}&\textbf{Attack}&\textbf{CNNSpot}\cite{CNNSpot}&\textbf{GenImage}\cite{GenImage}
  &\textbf{DRCT}\cite{DRCT}\\
    \midrule
    \multirow{4}{*}{FatFormer\cite{FatFormer}}
  &SimBA\cite{SimBA}&971.055&886.475& 960.372\\
  &Square\cite{Square}&\underline{178.402}&\underline{339.875}&\underline{213.075} \\
  &BruSLe\cite{BruSLe}&791.555 &555.690 &657.015 \\
    \cmidrule(r){2-2}\cmidrule(l){3-5}
  &\textbf{R$^2$BA}& \textbf{166.834}&\textbf{119.139}&\textbf{132.724} \\
    \bottomrule
    \end{tabular}
    \end{center}}
\end{table}
In practice, the number of queries to the detector is limited. A smaller number of queries can save time and control the budget. Therefore, as a black-box query-based attack, the number of queries is an important factor to consider. To ensure fairness, we check whether the adversarial point is reached after every 100 queries (we use 100 particles, with each particle representing 1 query). In Table \ref{tab:query}, we use FatFormer\cite{FatFormer} as the detector, which is the only one that covers all three datasets and generative models in its research and maintains good detection results for all three datasets. Evidently, the number of queries for R$^2$BA is significantly smaller than that for other black-box attacks. Therefore, compared to other query-based adversarial attacks, R$^2$BA is able to maintain excellent anti-detection performance with fewer queries.

\subsection{Effect of Particle Number}
To further explore the influence of the particle number on R$^2$BA optimization, we vary the number of particles and the number of iterations to compare the variations in ASR, BRISQUE\cite{BRISQUE}, SSIM\cite{SSIM}, and the number of queries under different particle numbers. To ensure the fairness of the experiments, the maximum number of queries is kept at 1000. 

\begin{figure}[!t]
    \centering
    \includegraphics[width=3.48in]{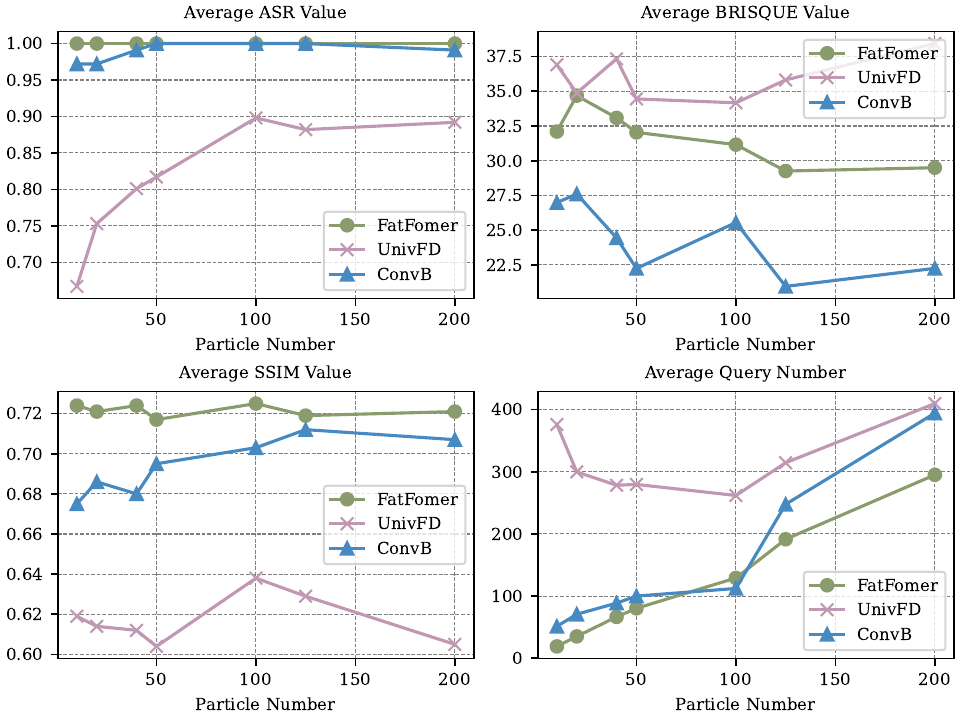}
    \caption{The effect of the number of particles on the attack success rate, BRISQUE, SSIM, and the number of queries.}
    \label{fig:particle}
\end{figure}

The experiments are shown in Fig. \ref{fig:particle}. We observe that when the number of particles increases, the ASR also increases. When the number of particles reaches 100, the ASR of all detectors reaches a relatively high level. However, the effect of the particle number on BRISQUE and SSIM is not significant. This indicates that when the intensity of post-processing is controlled within a certain range, the image quality of the adversarial examples generated by R$^2$BA does not fluctuate significantly with changes in other parameters of R$^2$BA.

Interestingly, in the UnivFD(DRCT) experiment, although we typically believe that an increase in the number of particles leads to a rise in the number of queries, the experimental results show that the number of queries is actually higher with a lower number of particles. Building on the previous experimental results regarding the attack success rate, we know that UnivFD(DRCT) is a detector that is harder to attack. This suggests that when attacking robust detectors, our optimization algorithm struggles to find a suitable direction for optimization with a lower number of particles, and thus more queries are needed for exploration. On the other hand, when the number of particles is higher, the query cost for a single iteration increases, and the increase in the query base leads to a multiplicative trend in the number of queries.

Therefore, choosing the right number of particles is crucial for optimizing the attack success rate of the algorithm. From the experimental results, it can be seen that choosing 100 particles and carrying out 10 iterations of the attack is the most ideal choice, as it strikes a better balance between the attack success rate and the number of queries. 

\subsection{Ablation Experiments}
\begin{table*}
    \setlength{\tabcolsep}{0.33mm}{
    \renewcommand\arraystretch{1.05}
    \caption{Ablation experiments on the optimization algorithm and post-processing techniques \label{tab:ablation}}
    \begin{center}
    \begin{tabular}{p{0.9cm}<{\centering} p{1.3cm}<{\centering} 
    p{0.9cm}<{\centering} p{0.9cm}<{\centering} p{0.9cm}<{\centering} p{0.9cm}<{\centering}
    p{0.9cm}<{\centering} p{0.9cm}<{\centering} p{0.9cm}<{\centering} p{0.9cm}<{\centering}
    p{0.9cm}<{\centering} p{0.9cm}<{\centering} p{0.9cm}<{\centering} p{0.9cm}<{\centering}
    p{0.9cm}<{\centering} p{0.9cm}<{\centering} p{0.9cm}<{\centering} p{0.9cm}<{\centering}
    }
    \toprule  
    \multirow{2}{*}{\textbf{Dataset}}& \multirow{2}{*}{\textbf{Attacks}}
  &\multicolumn{4}{c}{\textbf{FatFormer\cite{FatFormer}}}
  &\multicolumn{4}{c}{\textbf{UnivFD(DRCT)\cite{DRCT}}}
  &\multicolumn{4}{c}{\textbf{ConvB(DRCT)\cite{DRCT}}}
  &\multicolumn{4}{c}{\textbf{Average}}\\
    \cmidrule(lr){3-6}\cmidrule(lr){7-10}\cmidrule(lr){11-14}\cmidrule(l){15-18}
  &
  &ASR↑&BRIS↓&SSIM↑&Query↓
  &ASR↑&BRIS↓&SSIM↑&Query↓
  &ASR↑&BRIS↓&SSIM↑&Query↓
  &ASR↑&BRIS↓&SSIM↑&Query↓\\
    \cmidrule(){1-2}\cmidrule(lr){3-6}\cmidrule(lr){7-10}\cmidrule(lr){11-14}\cmidrule(l){15-18}
    \multirow{9}{*}{\rotatebox{90}{\textbf{GenImage}\cite{GenImage}}}
  & Random&0.244&33.186&0.711&N/A&0.005&31.284&0.726&N/A&0.292&31.915&0.725&N/A&0.181&32.128&0.721&N/A  \\
    \cmidrule(lr){2-2}\cmidrule(lr){3-6}\cmidrule(lr){7-10}\cmidrule(lr){11-14}\cmidrule(l){15-18}
  & w GB&0.781&68.481&0.708&322.5&0.789&73.279&0.699&298.4&0.704&51.660&0.767&453.7&0.758&64.473&0.725&358.2\\
  & w JPEG&\textbf{1.000}&36.360&0.878&\underline{131.2} &0.676&34.715&0.905&401.1&0.787&36.341&0.878&376.9&0.821&35.805&0.887&303.0\\
  & w GN& 0.538&24.743&0.883&576.9&0.249&23.233&0.884&805.4&0.602&22.807&0.868&564.8&0.463&23.594&0.878&649.0\\
  & w LS&0.144&24.394&0.959&898.8&0.141&23.469&0.965&886.5&0.315&24.193&0.943&813.9&0.200&24.019&0.956&866.4\\
    \cmidrule(lr){2-2}\cmidrule(lr){3-6}\cmidrule(lr){7-10}\cmidrule(lr){11-14}\cmidrule(l){15-18}
  & w/o GB&\textbf{1.000}&21.328&0.835&175.6&0.694&35.110&0.771&442.5&0.934&27.846&0.759&255.7&0.876&28.095&0.789&291.3\\
  & w/o JPEG&\underline{0.944} &23.841&0.693&203.1&0.876&30.768&0.643&\underline{272.6} &0.962&50.374&0.706&234.9&0.927&34.994&0.681&236.9\\
  & w/o GN&\textbf{1.000}&40.689&0.780&135.0&\underline{0.882} &44.357&0.701&272.8&\underline{0.991} &58.827&0.747&\underline{147.2} &\underline{0.957} &47.958&0.743&\underline{185.0}  \\
  & w/o LS&\textbf{1.000}&24.193&0.724&148.8&0.806&30.665&0.624&344.1&0.972&33.270&0.714&234.0&0.926&29.376&0.687&242.3\\
    \cmidrule(lr){2-2}\cmidrule(lr){3-6}\cmidrule(lr){7-10}\cmidrule(lr){11-14}\cmidrule(l){15-18}
  & \textbf{R$^2$BA}& \textbf{1.000}&25.209&0.827&\textbf{129.1}&\textbf{0.898}&32.098&0.665&\textbf{261.8}&\textbf{1.000}&19.494&0.804&\textbf{112.3}&\textbf{0.966}&25.600&0.765&\textbf{167.7} \\
    \bottomrule
    \end{tabular}
    \end{center}}
\end{table*}
    
To comprehensively validate the effectiveness of R$^2$BA, we conducted ablation experiments, as shown in Table \ref{tab:ablation}. The experiments are divided into three main parts:

{\textbf{Randomly generated post-processing intensity.}} We use randomly generated post-processing intensity (without optimization, so the query is N/A) to create adversarial examples for validating the effectiveness of R$^2$BA optimization. The results show that without guidance from the detector's fake probability, these examples struggle to cross decision boundaries and perform poorly in anti-detection. If the randomly generated intensity is weaker, the adversariality is also weaker, while stronger intensity will lead to severe distortion, causing the detector to classify the image as AI-generated as well.

{\textbf{R$^2$BA with single post-processing.}} Due to the different vulnerabilities of detectors to various post-processing techniques, there are performance differences between R$^2$BA with different post-processing. 
For example, the ASR of R$^2$BA with Gaussian noise is 0.249 on UnivFD(DRCT)\cite{DRCT}, but nearly 3 times higher on ConvB(DRCT)\cite{DRCT}. 
In addition, the anti-detection performance of R$^2$BA with JPEG compression is significantly better than that of R$^2$BA with light spot, likely because detectors are more sensitive to high-frequency information (which JPEG compression causes to be lost) than to changes in statistical features (which light spot alters). 

{\textbf{R$^2$BA without certain post-processing.}} The anti-detection performance of R$^2$BA without certain post-processing slightly decreases. This is because the effects of post-processing on features are not mutually exclusive. For example, JPEG compression leads to the loss of high-frequency information but also indirectly affects statistical features. Although optimization becomes harder, R$^2$BA without certain post-processing can still generate effective adversarial examples to achieve successful attacks, at the cost of more queries. As a result, R$^2$BA without certain post-processing shows lower anti-detection performance and requires more queries compared to R$^2$BA with all four post-processing.

Considering all metrics and comparing with ablated adversarial attack methods, R$^2$BA with 4 post-processing fusion attacks achieves the best results in both attack success rate and query count, while the image quality metrics BRISQUE\cite{BRISQUE} and SSIM\cite{SSIM} are either the best or close to the best.

\subsection{Attacks on Commercial AIGC Detectors}
\begin{table*}
    \setlength{\tabcolsep}{0.4mm}{
    \renewcommand\arraystretch{1.05}
    \caption{Attacks on Alibaba's Commercial AIGC Detection \label{tab:Alibaba}}
    \begin{center}
    \begin{tabular}{
        p{2cm}<{\centering} p{0.9cm}<{\centering} 
        p{0.9cm}<{\centering} p{0.9cm}<{\centering} p{0.9cm}<{\centering} 
        p{0.9cm}<{\centering} p{0.9cm}<{\centering} p{0.9cm}<{\centering} 
        p{0.9cm}<{\centering} p{0.9cm}<{\centering} p{0.9cm}<{\centering} 
        p{0.9cm}<{\centering} p{0.9cm}<{\centering} p{0.9cm}<{\centering} 
        p{0.9cm}<{\centering} p{0.9cm}<{\centering} p{0.9cm}<{\centering}}
    \toprule  
    \multirow{2}{*}{\textbf{Attacks}}
  &\multicolumn{4}{c}{\textbf{CNNSpot\cite{CNNSpot}}}
  &\multicolumn{4}{c}{\textbf{GenImage\cite{GenImage}}}
  &\multicolumn{4}{c}{\textbf{DRCT\cite{DRCT}}}
  &\multicolumn{4}{c}{\textbf{Average}}\\
  \cmidrule(lr){2-5}\cmidrule(lr){6-9}\cmidrule(lr){10-13}\cmidrule(l){14-17}
  & SR↓&BRIS↓&SSIM↑&Query↓&SR↓&BRIS↓&SSIM↑&Query↓&SR↓&BRIS↓&SSIM↑&Query↓&SR↓&BRIS↓&SSIM↑&Query↓\\
  \cmidrule(r){1-1}\cmidrule(lr){2-5}\cmidrule(lr){6-9}\cmidrule(lr){10-13}\cmidrule(l){14-17}
  Clean &1.000&N/A &N/A&N/A &1.000&N/A &N/A&N/A &1.000&N/A &N/A&N/A &1.000&N/A &N/A&N/A\\
  \cmidrule(r){1-1}\cmidrule(lr){2-5}\cmidrule(lr){6-9}\cmidrule(lr){10-13}\cmidrule(l){14-17}
  FGSM\cite{FGSM}&1.000&40.071&0.704&N/A &0.980&53.280&0.626&N/A &1.000&38.511&\underline{0.726}& N/A &0.993&\underline{43.954}& 0.685&N/A\\
  MIFGSM\cite{MIFGSM}&1.000&40.624&0.703&N/A &0.970&\underline{52.930}& 0.626&N/A &1.000&38.551&0.724&N/A &0.990&44.035&0.685&N/A\\
  PGD\cite{PGD} &1.000&40.543&0.703&N/A &0.970&53.008&0.625&N/A &1.000&\underline{38.384}& 0.723&N/A &0.990&43.978&0.684&N/A\\
  StatAT\cite{Evade}&0.230&\underline{39.148}& 0.602&N/A &\underline{0.165}& 54.794&0.552&N/A &0.135&40.615&0.609&N/A &0.177&44.852&0.587&N/A\\
  \cmidrule(r){1-1}\cmidrule(lr){2-5}\cmidrule(lr){6-9}\cmidrule(lr){10-13}\cmidrule(l){14-17}
  SimBA\cite{SimBA} &0.295&69.677&\underline{0.724}& 97.9&0.745&64.603&\underline{0.666}& 98.9&0.750&66.595&0.716&99.9&0.597&66.958&\underline{0.702}& 98.9\\
  Square\cite{Square} & \textbf{0.000} & 113.216 &0.535&\textbf{20.0} & \textbf{0.000} & 137.065 &0.457&\textbf{20.0} & \underline{0.005}& 112.051 &0.562&\underline{20.4}& \underline{0.002}& 120.777 &0.518&\underline{20.1}  \\
  BruSLe\cite{BruSLe}&\underline{0.185}& 90.584&0.578&\underline{45.9}& 0.505&86.623&0.486&\underline{68.7}& 0.440&77.847&0.569&66.8&0.377&85.018&0.544&60.5\\
  \cmidrule(r){1-1}\cmidrule(lr){2-5}\cmidrule(lr){6-9}\cmidrule(lr){10-13}\cmidrule(l){14-17}
  \textbf{R$^2$BA} & \textbf{0.000} & \textbf{25.593} & \textbf{0.740} & \textbf{20.0} & \textbf{0.000} & \textbf{25.410} & \textbf{0.691} & \textbf{20.0} & \textbf{0.000} & \textbf{29.640} & \textbf{0.733} & \textbf{20.0} & \textbf{0.000} & \textbf{26.881} & \textbf{0.722} & \textbf{20.0}\\
    \bottomrule
    \end{tabular}
    \end{center}}
\end{table*}
We use Alibaba's commercial AIGC detection API\cite{Alibaba} to compare the effectiveness of the attacks under real-world conditions. For budget and fairness, we set the maximum query to 100, the number of particles in R$^2$BA to 10, and check every 20 queries to determine whether the attack succeeds. For white-box attacks, we test adversarial examples generated by attacking the UnivFD\cite{UnivFD} and UnivFD(DRCT)\cite{DRCT} detectors, the two most robust detectors across three datasets.

\begin{figure}[!t]
    \centering
    \includegraphics[width=3.48in]{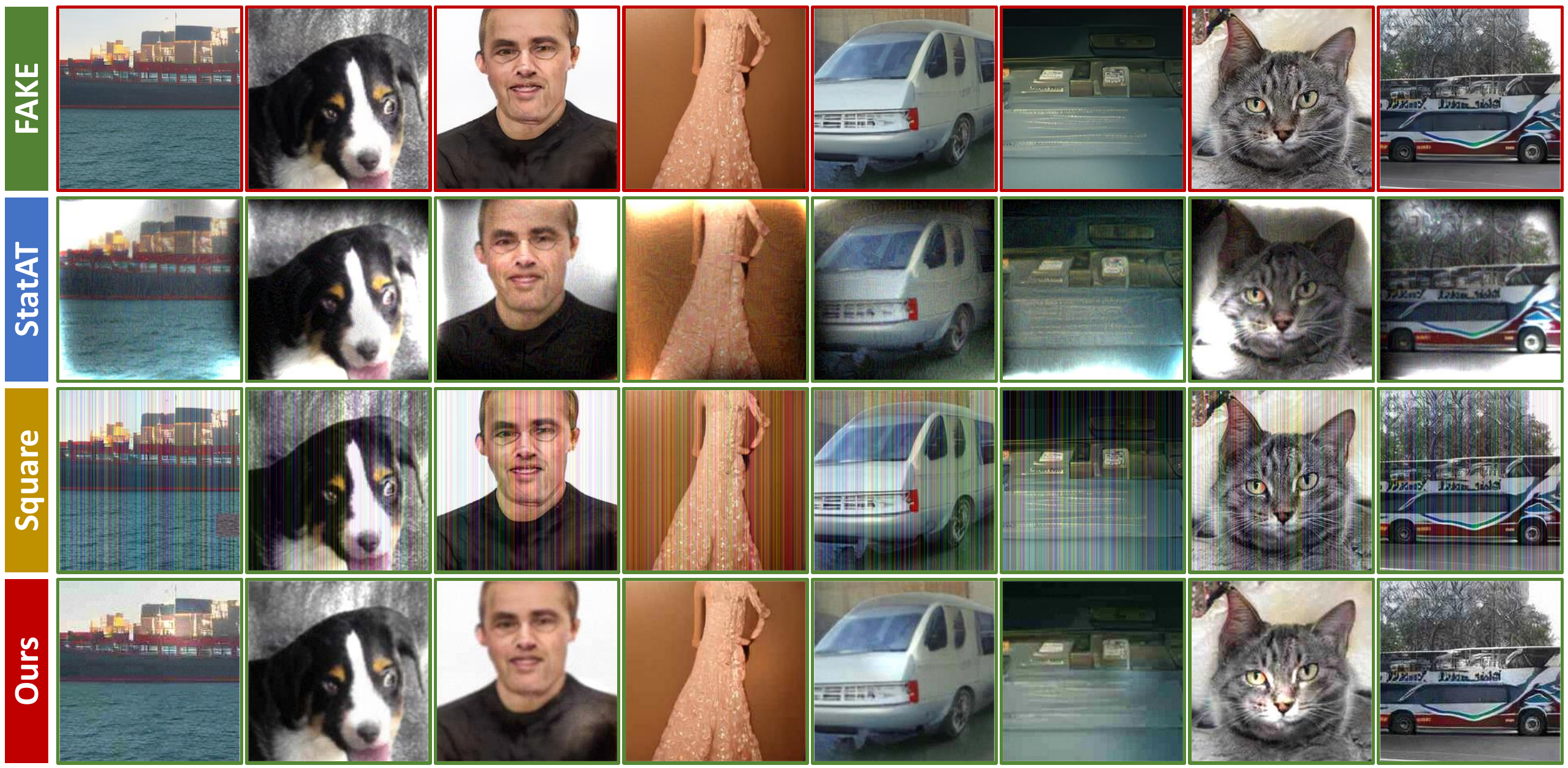}
    \caption{Visual comparison on Alibaba's commercial AIGC detection. The red box indicates AIGC and the green box indicates UGC. The labels for AIGC and UGC come from Alibaba AIGC detection API\cite{Alibaba}.}
    \label{fig:Alibaba}
\end{figure}

In Table \ref{tab:Alibaba}, \textit{clean} refers to the detection success rate (SR) of clean AIGC images, which is 100\% for all cases. N/A in \textit{clean} means irrelevant, and in white-box, N/A means query not applicable. Furthermore, in Fig. \ref{fig:Alibaba}, we present a visual comparison of two high-performance attacks\cite{Evade},\cite{Square}. Overall, the experiments show that RBA exhibits excellent anti-detection performance, high invisibility, and outstanding query efficiency in real-world scenarios. The experiment further demonstrates that AIGC detectors can classify post-processed AIGC images as User-generated images. 

\section{Conclusion}\label{section:Conclusion}
In this paper, we explore the robustness of GAN-based and diffusion-based AIGC detectors under adversarial example attacks. Given the uncertainty of detectors in real-world scenarios and the discovery that detectors vary in vulnerability to different types of post-processing, we propose a realistic-like robust black-box adversarial attack method, R$^2$BA, based on post-processing fusion optimization. Our method generates adversarial examples that are both robust and maintain high image quality by optimizing post-processing intensity, effectively evading both GAN-based and diffusion-based detection. We conduct extensive experiments on three GAN-based and diffusion-based AIGC datasets, as well as four GAN-based AIGC detectors, three diffusion-based AIGC detectors and one commercial AIGC detection API. The comprehensive experimental results demonstrate that the proposed attack method achieves promising anti-detection performance in both GAN-based and diffusion-based settings, under the original, robust, and real-world scenarios, respectively.

\bibliographystyle{IEEEtran}
\bibliography{reference.bib}
\end{document}